\documentclass[final,12pt]{elsarticle}

\usepackage{lineno,hyperref}
\usepackage{amsmath}
\usepackage{amssymb}
\DeclareMathOperator{\Tr}{Tr}

\usepackage{tikz} % for engine model
\usepackage{verbatim}
\usetikzlibrary{arrows,shapes,fit,scopes,calc,matrix,positioning,decorations.pathmorphing}
\usepackage{subfig}
\usepackage[short]{optidef}
\usepackage{todonotes}

\graphicspath{{Figures/}}

\newcommand{\Fig}{Figure~}

\newtheorem{defn}{Definition}

\journal{Control Engineering Practice}

\begin{document}

\begin{frontmatter}

%% Title, authors and addresses

%% use the tnoteref command within \title for footnotes;
%% use the tnotetext command for theassociated footnote;
%% use the fnref command within \author or \address for footnotes;
%% use the fntext command for theassociated footnote;
%% use the corref command within \author for corresponding author footnotes;
%% use the cortext command for theassociated footnote;
%% use the ead command for the email address,
%% and the form \ead[url] for the home page:
%% \title{Title\tnoteref{label1}}
%% \tnotetext[label1]{}
%% \author{Name\corref{cor1}\fnref{label2}}
%% \ead{email address}
%% \ead[url]{home page}
%% \fntext[label2]{}
%% \cortext[cor1]{}
%% \affiliation{organization={},
%%             addressline={},
%%             city={},
%%             postcode={},
%%             state={},
%%             country={}}
%% \fntext[label3]{}

\title{Data-Driven Fault Diagnosis Analysis and Open-Set Classification of Time-Series Data}
\tnotetext[mytitlenote]{Conflict of interest - none declared.}

%% use optional labels to link authors explicitly to addresses:
%% \author[label1,label2]{}
%% \affiliation[label1]{organization={},
%%             addressline={},
%%             city={},
%%             postcode={},
%%             state={},
%%             country={}}
%%
%% \affiliation[label2]{organization={},
%%             addressline={},
%%             city={},
%%             postcode={},
%%             state={},
%%             country={}}

\author{Andreas Lundgren}\ead{andreas-lundgren@live.se}
\author{Daniel Jung\corref{mycorrespondingauthor}}\ead{daniel.jung@liu.se}
\address{Department of Electrical Engineering, Link\"{o}ping University, Link\"{o}ping, Sweden}  % Please supply    
\cortext[mycorrespondingauthor]{Corresponding author}       
                         
\begin{abstract}
Fault diagnosis of dynamic systems is done by detecting changes in time-series data, for example residuals, 
caused by system degradation and faulty components. The use of general-purpose multi-class classification methods for fault 
diagnosis is complicated by imbalanced training data and unknown fault classes. Another complicating factor 
is that different fault classes can result in similar residual outputs, especially for small faults, which causes 
classification ambiguities. In this work, a framework for data-driven analysis and open-set classification is 
developed for fault diagnosis applications using the Kullback-Leibler divergence. A data-driven fault 
classification algorithm is proposed which can handle imbalanced datasets, class overlapping, and unknown faults. In addition, 
an algorithm is proposed to estimate the size of the fault when training data contains information from known fault realizations. 
An advantage of the proposed framework is that it can also be used for quantitative analysis of fault diagnosis 
performance, for example, to analyze how easy it is to classify faults of different magnitudes. To evaluate the 
usefulness of the proposed methods, multiple datasets from different fault scenarios have been collected from 
an internal combustion engine test bench to illustrate the design process of a data-driven diagnosis system, 
including quantitative fault diagnosis analysis and evaluation of the developed open set fault classification algorithm. 
\end{abstract}

%%Research highlights
\begin{highlights}
\item A data-driven framework for fault diagnosis is proposed which is used for quantitative analysis, open-set classification, and fault size estimation.  
\item Fault classes are modeled using the Kullback-Leibler divergence where batches of data are classified by comparing their distribution to training data.
\item The proposed framework is validated on a set of residual generators using real data from an engine test bench operating during different faulty conditions.
\end{highlights}

\begin{keyword}
Open-set classification \sep Fault diagnosis \sep Fault estimation \sep Kullback-Leibler divergence \sep Machine learning.
%% keywords here, in the form: keyword \sep keyword

%% PACS codes here, in the form: \PACS code \sep code

%% MSC codes here, in the form: \MSC code \sep code
%% or \MSC[2008] code \sep code (2000 is the default)

\end{keyword}

\end{frontmatter}

%% \linenumbers

%% main text
\section{Introduction}

Fault diagnosis of technical systems deals with the problem of detecting and isolating faults 
by comparing model predictions of nominal system behavior and data from sensors
mounted on the monitored system \cite{jung2018combining}. Early detection 
of faults and identifying their root cause are important to improve system 
reliability and to be able to select suitable counter measures. 
Two common approaches of fault diagnosis are model-based and data-driven \cite{gao2015survey}. 

Model-based fault diagnosis relies on a mathematical model describing the nominal 
system behavior, where the model is derived based on physical insights about the system. 
Residuals are computed by comparing model predictions and sensor data to detect 
inconsistencies caused by faults \cite{isermann2005model,jiang2020optimized}. 
Data-driven fault diagnosis uses training data from different operating conditions and 
faulty scenarios to capture the relationship between a set of input and 
output signals \cite{dai2013model}. The output signal could be a feature or sensor value to be 
predicted, which is referred to as regression, or the class label that input 
data belong to, referred to as classification \cite{hastie2009elements}.
 
Fault detection and isolation are complicated by prediction inaccuracies 
and measurement noise \cite{eriksson2013method}. For nonlinear dynamic systems, 
the sensor data distribution varies due to different operating conditions which requires 
complex data-driven classifiers to capture the distribution of different classes to 
distinguish between faults. One solution, see for example \cite{jung2020residual}, is to use 
residual generators to compute residual outputs as features to filter out the system dynamics 
before classifying faults. 

Even though system dynamics can be filtered out in feature data, the distribution of faulty 
data is not only dependent on fault class but also the realization of the fault, e.g., fault magnitude 
and excitation. Thus, collecting representative training data from various fault scenarios that 
can occur in the system is a complicated task, especially when developing a diagnosis system 
during early system life when failures are still rare \cite{theissler2017detecting}. 
General-purpose models for supervised multi-class classification, such as 
Random Forests and Neural Networks, have a risk of misclassifications 
when training data are limited since these methods assume that training data 
are representative of all data classes. In many technical applications, it is 
necessary that a diagnosis system can handle limited, and imbalanced, training data
and is able to detect likely unknown fault scenarios that would require special attention by an operator 
or technician \cite{theissler2017detecting}. One application is, for example, computer aided 
troubleshooting where a priority list of plausible fault hypotheses can guide a technician when 
identifying the faulty component \cite{pernestaal2012modeling}. It is also important to update the models
over time as more training data become available to improve classification performance 
\cite{jung2018combining,sankavaram2015incremental,dong2017method}. 

\begin{figure}[h!]
	\centering
	\includegraphics[width=\linewidth]{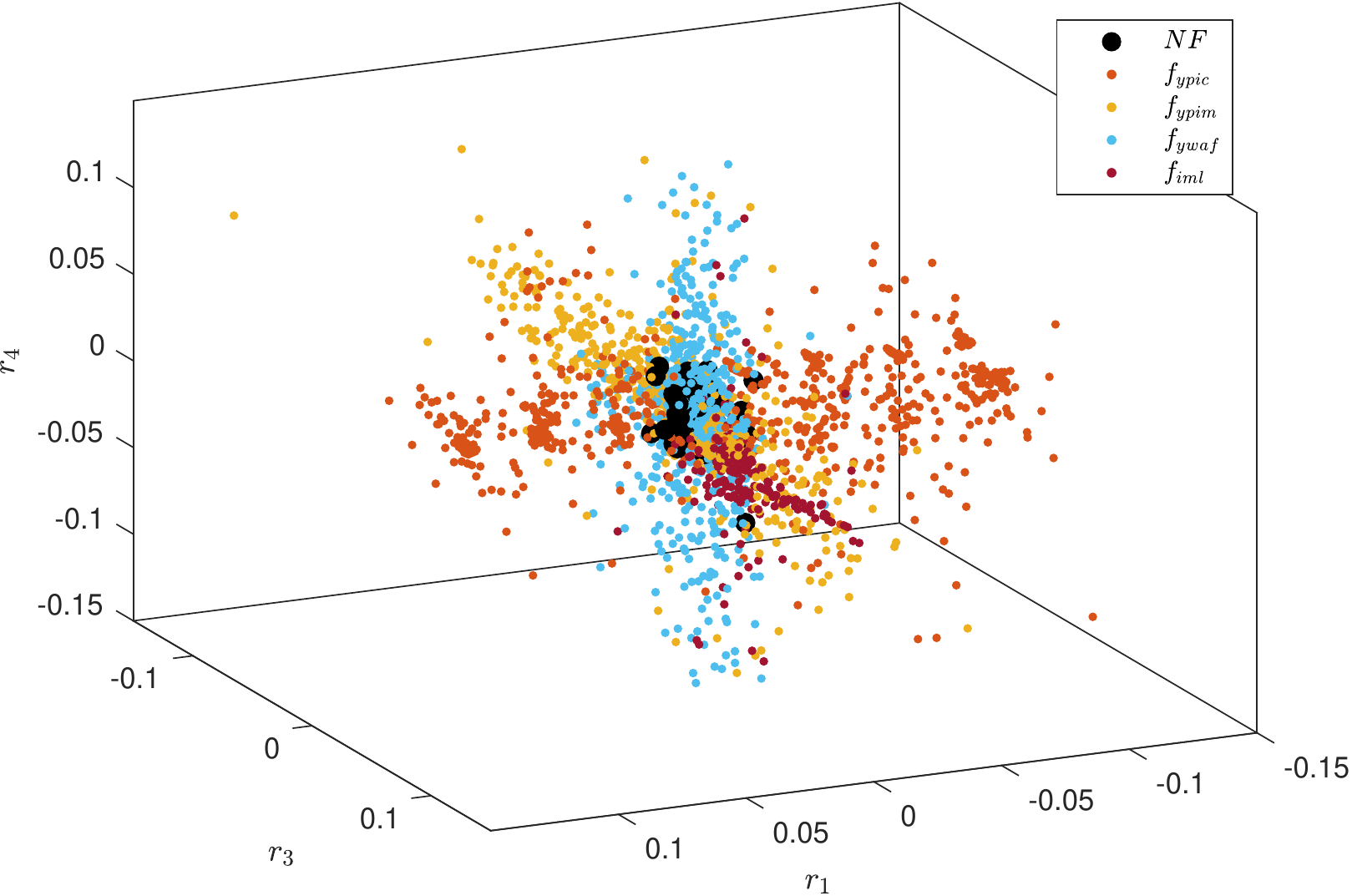}
	\caption{The figure shows residual data from the engine case study where data from 
	three of four residual generators are plotted against each other. The different colors 
	correspond to data from different fault classes where $NF$ represents the fault-free class.}
	\label{fig:residual_faults}
\end{figure}

Another complicating factor of data-driven fault diagnosis is class overlapping \cite{lee2018overlap} 
when different faults have similar impact on the system behavior. One example is when trying to classify 
small faults at an early stage, which can result in classification ambiguities \cite{campagner2020three}. 
This is illustrated in \Fig\ref{fig:residual_faults} where real data from a set of 
residual based fault detectors for an internal combustion engine test bench \cite{jung2020residual}, 
are plotted against each other. The figure shows that fault-free data (No Fault - $NF$) and data from the different fault 
classes $f_i$ are overlapping, especially data from small faults close to the origin. 
It can also be different fault classes that have similar impact on system behavior, e.g., the faults $f_{ypim}$ and $f_{yiml}$ 
that are both related to the engine intake manifold. Thus, it is not desirable to compute only the most likely fault 
hypothesis (class label), since this could fail to identify the true fault, but instead 
find all plausible fault hypotheses that can explain the observed data. 
Note that this is different from multi-label classification since each sample belongs to one 
fault class but can be explained by multiple classes \cite{campagner2020three}. 

Different classifiers that can handle overlapping classes have been proposed in, for example, 
\cite{lee2018overlap}. Because of overlapping fault classes, analysis methods that can quantify how 
easy it is to distinguish between different fault classes are necessary during the diagnosis system 
design process \cite{eriksson2013method}. Applying quantitative analysis early during the 
system development phase can be used to predict if, for example, fault diagnosis performance 
requirements can be met. Some of the most common performance analysis methods of multi-class 
classifiers are cross-validation and confusion matrices, see e.g. \cite{hastie2009elements}. Note that these methods are used 
to evaluate the performance of a given classifier rather than analyzing the distribution of data from different 
fault classes. There are also methods, such as \cite{van2008visualizing}, that can help to visually 
analyze multi-dimensional data. However, it is not obvious how to use this information to quantitatively 
measure separation between different classes.

Besides fault detection and isolation, an important task of diagnosis systems is to track system 
degradation, for example by continuously estimating the size, or severity, of a fault that is present 
in the system. Accurate information about fault size is important for predictive maintenance 
\cite{larsson2014gas}, prognostics \cite{daigle2012comparison}, and fault reconstruction 
algorithms \cite{yan2007nonlinear}. 
Several model-based techniques for fault size estimation have been proposed, see e.g. 
\cite{daigle2012comparison}. Still, it is a non-trivial problem to estimate the fault size when no fault models are available. 

The main objective of this work is to develop a framework for data-driven analysis and 
classification for fault diagnosis of dynamic systems. The first contribution is a quantitative analysis method 
for data-driven fault classification using the Kullback-Leibler (KL) divergence to 
model data from different fault classes and to analyze fault detection and isolation performance 
for a given set of features. The second contribution is a data-driven open set classification algorithm 
designed for fault diagnosis applications that can handle: limited training data, unknown fault 
classes, and overlapping regions of different classes. A third contribution is a data-driven fault size estimation 
algorithm, using the same modeling framework, when training data are available with known fault 
sizes. As a case study, real data are collected from an internal combustion engine test bench which is 
operated during both nominal and faulty operation \cite{jung2018combining}. 

The outline of this paper is as follows. First, the problem statement is presented 
is Section~\ref{sec:problemformulation}. Related research is summarized in 
Section~\ref{sec:relatedresearch} and some background to fault diagnosis and 
quantitative fault detection and isolation analysis is given in 
Section~\ref{sec:background}. Then, the proposed framework for data-driven 
quantitative analysis is presented in Section~\ref{sec:modeling}, the proposed 
open set fault classification algorithm in Section~\ref{sec:classification}, and the 
fault size estimation algorithm in Section~\ref{sec:estimation}. The internal combustion 
engine case study 
is described in Section~\ref{sec:casestudy} and the results of the experiments are presented in 
Section~\ref{sec:evaluation}. Finally, some conclusions are summarized in Section~\ref{sec:conclusions}.     
\section{Problem Statement}
\label{sec:problemformulation}

The main objective is to develop a data-driven framework for supervised machine learning and 
data-driven analysis to systematically address the complicating factors of data-driven fault diagnosis 
of non-linear dynamic systems. The characterizing properties of the data-driven fault diagnosis problem are 
summarized in the following bullets:
\begin{itemize}
\item Training data are not representative of all relevant fault realizations 
\item There are both known and unknown fault classes
\item Feature data from different fault classes are overlapping
\item The distribution of feature data is not only depending on fault class, but also fault magnitude and excitation  
\end{itemize} 
Even though there has been significant work done in data-driven fault diagnosis and machine learning, 
it is still an open question how to address all these complicating factors in a common framework. 

In the problem formulation, it is assumed that a set of $n$ features has been developed to monitor 
a non-linear dynamic system. The set of computed features is time-series data, for example residuals 
computed from sensor outputs, that can be used to detect faults in the system. It is assumed that 
a set of training data has been collected from different nominal and faulty scenarios where the 
samples are correctly labelled. Labelling mainly considers which fault class each sample belongs to, 
but it will also be investigated what can be done when information about fault sizes is available as well. 
It is assumed that the available training data fulfill the characterizing properties listed above.

The first objective of this work is to develop a data-driven framework for modeling of different fault 
classes. To analyze the properties of training data, a quantitative method will be developed that is 
able to analyze the ability to classify the different faults. The proposed method should be able to 
quantify how easy it is to distinguish between different fault classes which can give valuable 
insights about the nature of different faults.  

The second objective is to develop a fault classification algorithm that is designed to detect 
and classify known fault classes that can explain the observations but also identify scenarios with 
unknown faults. Because feature data is assumed to overlap between different classes, the classifier 
should be designed such that the computed fault hypotheses are based on the characterizing 
properties of training data to avoid misclassifications.  

The third objective considers the case when training data are collected from different fault realizations with 
known fault magnitudes, e.g., a given sensor bias or leakage diameter. The purpose is to use the proposed 
framework to estimate the magnitude of a detected known fault to track the system health when no 
mathematical model is available to estimate the fault. 
\section{Related Research}
\label{sec:relatedresearch}

There has been an increasing focus on data-driven classification algorithms that can handle 
unknown classes, which is referred to as open set classification, see e.g. \cite{scheirer2013toward}, 
and overlapping classes, see e.g. \cite{lee2018overlap}. However, it is still an open problem 
how to address all these mentioned characteristics of data-driven fault diagnosis problems.
Open set classification has been used in, e.g., computer vision, to deal with unknown 
classes not covered by training data \cite{scheirer2013toward}. 
Different algorithms have been developed to solve the open set classification problem, 
for example Weibull-calibrated support vector machines \cite{scheirer2014probability} 
and extreme value machines \cite{rudd2018extreme}. 

Different data-driven approaches have been 
proposed for open set fault classification to handle both known and unknown fault scenarios, 
for example, one-class support vector machines \cite{jung2018combining,jung2020data}, conditional 
Gaussian network \cite{atoui2019single}, ensemble methods \cite{theissler2017detecting}, and 
Hidden Markov Models \cite{yan2018fault}. Several papers have proposed open set classification 
algorithms for machinery condition monitoring, see for example \cite{tian2018subspace,wang2021novel,yu2021deep}.  
The authors of \cite{tian2018subspace} propose a hybrid approach combining different data-driven 
methods and subspace learning. 
In \cite{wang2021novel}, a neural network with residual learning blocks is proposed and \cite{yu2021deep} 
proposes a deep learning neural networks-based approach to handle the situations where training 
data and test data are collected from different operating conditions. 
In \cite{michau2019domain}, a domain adaptation open set classification approach is proposed 
combining auto-encoders, a proposed homothety loss, and an origin discriminator, to monitor a 
system using training data from other similar systems. 
With respect to previous works, the proposed data-driven framework can also be used to evaluate 
fault diagnosis performance given a set of features. The same framework is used to design an open set fault 
classification algorithm that handles overlapping classes, including scenarios with unknown faults, 
and estimates the fault sizes of known fault classes by evaluating the probability distribution of 
feature data.  

Quantitative fault detection and isolation analysis has been considered in, for example, 
\cite{eriksson2013method,li2020gap}. In \cite{eriksson2013method}, the KL divergence 
is used to analyze time-discrete linear descriptor models. Similar approaches have also been used 
in e.g. \cite{fu2020data} and \cite{gienger2020robust}. In \cite{li2020gap}, diagnosis performance is 
measured based on the distance between different kernel subspaces. One application of 
quantitative analysis is sensor selection, see for example \cite{jung2020sensor,jiang2019multi}. 
The KL divergence has also been used for fault detection, see for example \cite{yan2018fault,chen2018improved}. 
With respect to these mentioned works, a data-driven framework is proposed here for quantitative 
fault diagnosis performance analysis, open set fault classification, and fault size estimation.  

Developing a mathematical model of complex systems with sufficient accuracy 
for fault diagnosis purposes is a time-consuming process \cite{naderi2017data,shen2020hybrid}. 
This has motivated the use of machine learning and data-driven fault diagnosis 
methods to instead learn the system behavior from collected data. Still, 
model-based techniques can be used to derive useful features for system 
monitoring, see e.g. \cite{larsson2014gas}. One example is model-based 
residuals that can filter out system dynamics and improve signal fault-to-noise ratio 
which reduces the need for complex data-driven classification models to distinguish 
between different fault classes \cite{jung2018combining}. 

Several recent papers consider hybrid diagnosis system designs for dynamic systems 
combining, e.g., model based residual generators and machine learning. In 
\cite{luo2009integrated}, model-based residuals and observers are used for fault detection in 
an automotive braking system where a hybrid fault isolation approach is proposed combining 
model-based fault isolation and support vector machines. The same braking system case study 
is also considered in \cite{slimani2018fusion} where an ensemble classifier is proposed combining 
both model-based and data-driven fault detectors. In \cite{khorasgani2018methodology}, both 
sensor data and residual data are used as input to a tree augmented naive Bayes fault classifier. 
In \cite{zhang2019knowledge}, feature selection using neural networks is applied before training 
the fault classifiers. In \cite{tidriri2018generic}, model-based residuals and sensor data are used 
as inputs to a Bayesian network to perform fault classification and in \cite{matei2018classification} 
model data features are extracted and fed into a neural network classifier. With respect 
to previous work, the proposed diagnosis system can identify unknown fault scenarios 
and estimate the fault size of known faults.

Fault size estimation is important for tracking system degradation \cite{larsson2014gas}, prognostics 
\cite{daigle2012comparison} and fault reconstruction \cite{yan2007nonlinear}. 
In \cite{wan2016data}, a robust linear receding horizon model-based approach is proposed for 
fault estimation for linear discrete-time state space models with additive faults. In \cite{naderi2018data},
a data-driven fault estimation algorithm is evaluated on an aircraft gas turbine case study 
where a linear state-space model, is estimated from data, and then used to estimate the fault size 
by assuming additive fault models. In these mentioned works, the faults are included as parameters 
in the system model and estimated using model-based techniques. With respect to these works, a 
data-driven fault size estimation algorithm is proposed when no model of the fault is available.  

Some work on data-driven fault size estimation has been done, see for 
example \cite{Sawalhi2007,Guo2016}. In \cite{Sawalhi2007}, faults are assumed to 
appear as pulses in the time domain data which is inherently tied to the bearing case. 
In \cite{Guo2016}, Paris' formula \cite{Paris1963}, estimating crack growth in bearings, is 
used to interpolate between distributions from known fault sizes. A hybrid method is
proposed in \cite{ezzat2020model} for structural health monitoring combining finite 
element models and machine learning to estimate fault location and severity. In \cite{zhang2020incipient}, 
a fault detection and estimation scheme are proposed for incipient faults using the Jensen-Shannon 
divergence measure. With respect to previous work, fault size estimation is formulated as an 
optimization problem using the KL divergence and training data from different types of fault realizations. 
\section{Background}
\label{sec:background}
To address the fault diagnosis problem, a data-driven framework 
is proposed to model data from different fault classes and 
quantify fault diagnosis performance. A summary of the relevant results and 
definitions from previous work \cite{eriksson2013method} is presented here.

\subsection{Modeling Fault Classes} \label{sec:fault_modeling}

\begin{figure}[h!]
	\centering
	\includegraphics[width=\linewidth]{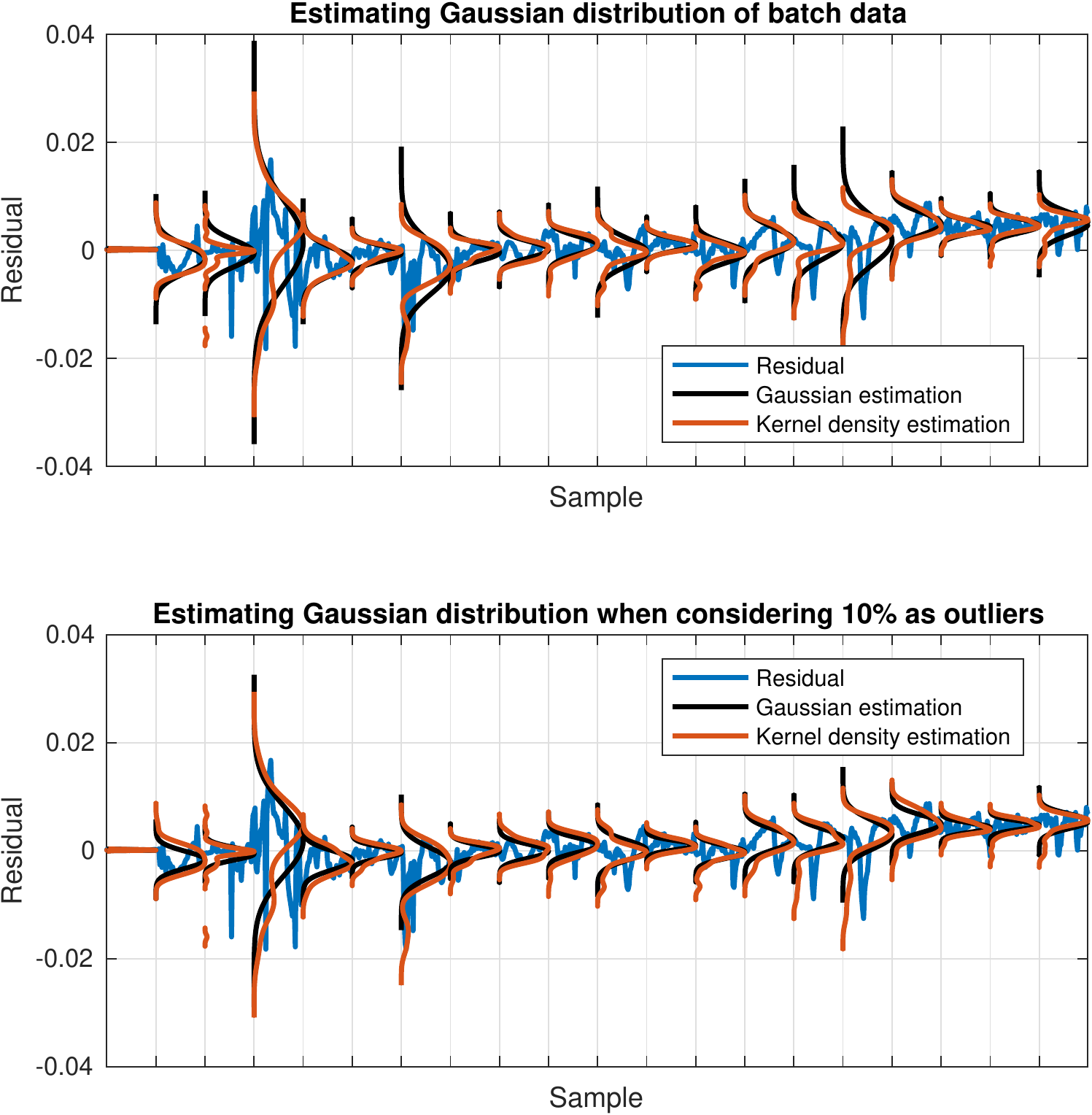}
	\caption{An illustration of the one-dimensional residual which is partitioned into consecutive batches and the estimated data distribution of each batch. The distribution is estimated using a Gaussian distribution and a kernel density estimation. The upper plot shows the estimated Gaussian distribution using all samples in each batch and the lower plot the estimated distribution after removing 10\% of the outliers in each batch.}
	\label{fig:batch_pdfs}
\end{figure}

Let $\bar{r} = (r_{1}, r_{2}, \ldots, r_{n})$ denote a set of $n$ 
signals or features, for example residuals. A sample of $\bar{r}$ at time index $t$, 
denoted $\bar{r}_t$, belongs to one of $m$ known fault classes $\{f_1, f_2, \ldots, f_m\}$. 
The fault-free class is denoted $NF$ (No Fault). 
To capture the impact of model uncertainties and measurement noise, each dataset 
$\{\bar{r}_1, \bar{r}_2, \bar{r}_3, \ldots\}$ is partitioned into batches of size $N$, for example $R = \{\bar{r}_{t-N+1}, \bar{r}_{t-N+2}, \ldots, \bar{r}_{t}\}$, 
where the distribution of data in each batch is modeled as a probability density function (pdf) $p = p(R)$. An illustration 
of the one-dimensional case is shown in \Fig\ref{fig:batch_pdfs}. The figure shows partitioned 
time-series data and the estimated Gaussian distribution for each batch. For comparison, the distribution 
for each batch is also estimated using a kernel density estimator. It is assumed that all samples in one batch belong to 
the same fault class.   

The pdf $p(R)$ of each batch $R$ varies depending on different 
system operating conditions, such as, operating point and fault realization. Let $\Omega_i 
= \Omega_i(\bar{r})$ denote the set of pdfs, based on the feature set $\bar{r}$, that can be explained by fault $f_i$ where 
$p(R) \in \Omega_i(\bar{r})$ is used to denote one pdf $p(R)$ in the set. 
For example, if $p \sim \mathcal{N}(\mu, \Sigma)$ is a multivariate normal distribution, the 
mean estimate $\hat{\mu}$ and covariance estimate $\hat{\Sigma}$ can be obtained from a batch of $N$ samples, $R = \{\bar{r}_1, \bar{r}_2, \ldots, \bar{r}_N\}$, as
\begin{equation}
\hat{\mu} = \frac{1}{N}\sum_{t=1}^N \bar{r}_t \quad \quad \hat{\Sigma} = \frac{1}{N-1}\sum_{t=1}^N (\bar{r}_t - \hat{\mu})(\bar{r}_t - \hat{\mu})^{\text{T}}
\end{equation}
 
The following definition is used to model each fault class. 
\begin{defn}[Fault mode]
Let $\{\bar{r}_1, \bar{r}_2, \bar{r}_3, \ldots\}$ be a set of time series data, when fault $f_i$ is present, that is 
partitioned into a set of consecutive batches where each batch is represented by a pdf $p(R)$. 
A fault mode is defined by the set $\Omega_i(\bar{r})$ 
with corresponding pdfs $p(R) \in \Omega_i(\bar{r})$ for each batch that can be explained by fault class $f_i$.
\end{defn}  
To simplify notation, $p$ and $\Omega_i$ are used where the dependence on 
$R$ and $\bar{r}$, respectively, are omitted. Different fault classes $f_i$ are represented by different modes 
$\Omega_i$ where $\Omega_{NF}$ is used to denote the fault-free mode. 
Note that each fault mode $\Omega_i$ is modeled independently and 
there can be pdfs that belong to multiple fault modes, i.e., different fault realizations can result 
in the same distribution of data in one batch.

Detectability and isolability of different fault classes depend on if there are 
observations (pdfs) that can be explained by one fault class but not another, i.e., a fault 
class $f_i$ can be isolated from another fault $f_j$ if there is a $p \in \Omega_i$ 
such that $p \not\in \Omega_j$. 
\begin{defn}[Fault isolability]
A fault class $f_i$ is isolable from another fault class $f_j$ if 
$\Omega_i\setminus\Omega_j \neq \emptyset$. If a fault $f_i$ is isolable from the 
fault-free class, then the fault is said to be detectable.
\label{defn:isolability}
\end{defn}
Using Definition~\ref{defn:isolability}, it is possible to analyze fault detection and isolation 
performance by comparing the different sets $\Omega_i$ where $i = 1,2, \ldots, m$ \cite{jung2015analysis}. 
Note that, even though fault modes are isolable from each other it does not mean that 
all pdfs $p \in \Omega_i$ can distinguish $f_i$ from $f_j$. These ambiguities are caused by, for example, 
model uncertainties and sensor noise or lack of fault excitation. One illustration of fault excitation is leakage 
detection where the flow through the orifice, and thus detection performance, depends on the pressure 
difference since a small difference will result in a small leakage flow which can be difficult to detect. 

\subsection{Fault Classification by Rejection of Fault Hypotheses}   

Assuming that faults can be small or have varying impact on the feature set $\bar{r}$, 
each fault class $f_j$ is modeled such that the nominal mode
$\Omega_{NF} \subseteq \Omega_j$. An implication is that it is possible to distinguish 
faults from nominal behavior but not vice versa. This is consistent with the principles of consistency-based 
fault isolation algorithms, such as \cite{de1987diagnosing}, where $p \in \Omega_{NF}$ 
can be explained by a fault free system but also that 
the system is faulty (in case the fault has not yet been detected). 

Instead of selecting the 
most likely target class, the set of plausible fault hypotheses is computed by rejecting fault classes that 
cannot explain data. A fault is detected when the fault-free class is rejected, i.e., when 
$p \not\in \Omega_{NF}$. Similarly, fault classification is not performed by selecting the 
most likely fault class but by rejecting fault hypotheses, e.g., a fault class $f_j$ is rejected if 
$p \not\in \Omega_{\text{j}}$. This principle will be used here when performing data-driven fault isolation.

\subsection{Quantitative Fault Diagnosis Analysis}

The similarities between the different fault modes $\{\Omega_1, \Omega_2, \ldots, \Omega_m\}$ 
can be used to analyze fault diagnosis performance for a given system. However, 
only analyzing qualitative performance, such as fault isolability in Definition~\ref{defn:isolability}, 
does not give sufficient information regarding how easy it is to detect and isolate different faults. 
If feature distributions are used to model each fault class, one way to quantify fault diagnosis 
performance is to use the KL divergence to measure the similarity between two pdfs \cite{eriksson2013method}. 
The KL divergence can be used as a similarity measure between pdfs 
and is defined as \cite{kullback1951information}
\begin{equation}
K(p\|q) = \int p \log \left( \frac{p}{q}\right) dp = \mathbb{E}_p\left[ \log \frac{p}{q}\right]
\label{eq:kl}
\end{equation}
where $\mathbb{E}_p\left[\cdot\right]$ denotes the expected value given the pdf $p$.

From a fault diagnosis perspective, \eqref{eq:kl} can be interpreted as the 
expected value of a log-likelihood ratio test determining if $\bar{r}$ is drawn from a 
distribution with a pdf $p$ or $q$ when $p$ is the true density function. If $p$ and $q$ are two pdfs 
representing two different fault realizations, the larger the value of $K(p\|q)$
the easier it is to distinguish $p$ from $q$ when $p$ is true. However, since each fault can 
have different realizations, quantitative isolation performance of fault $f_i$ 
with realization $p$ from another fault $f_j$ is defined by the smallest value of $K(p\|q)$ for all 
$q \in \Omega_j$. This measure is proposed in \cite{eriksson2013method}, 
called distinguishability, and is defined as
\begin{equation}
\mathcal{D}_{i,j}^*(p) = \min_{q \in \Omega_j} K(p \| q)
\label{eq:dij_true}
\end{equation}
If the distribution of batch $\bar{r}$ is described by 
pdf $p$ when fault $f_i$ is present, i.e., $p \in \Omega_i$, the distinguishability 
measure $\mathcal{D}_{i,j}^*(p)$ quantifies how easy it is to isolate 
$f_i$ from $f_j$. A large value of \eqref{eq:dij_true} corresponds to an easier isolation 
problem \cite{jung2020sensor}. Fault detection performance 
is denoted $\mathcal{D}_{i,NF}^*(p)$. The distinguishability measure 
$\mathcal{D}_{i,j}^*(p)$ is non-negative and equal to zero if and only if 
$p \in \Omega_j$, i.e., when it is not possible to isolate from fault class $f_j$. 
Another property of the distinguishability measure is that if 
$\Omega_{NF} \subseteq \Omega_{j}$, then \cite{eriksson2013method}
\begin{equation}
\mathcal{D}_{i,NF}^*(p) \geq \mathcal{D}_{i,j}^*(p)
\label{eq:dij_det_isol}
\end{equation}
This result can be interpreted as that it is easier to detect a fault $f_i$ than to 
isolate it from another fault $f_j$.

If $p \sim \mathcal{N}(\mu_p, \Sigma_p)$ and 
$q \sim \mathcal{N}(\mu_q, \Sigma_q)$ are two $n$-dimensional 
multivariate normal distributions with known mean vectors, $\mu_p, \mu_q \in \mathbb{R}^n$, and 
covariance matrices, $\Sigma_p, \Sigma_q \in \mathbb{R}^{n \times n}$, 
$K(p \| q)$ can be computed analytically as \cite{bishop2006pattern}
\begin{equation} \label{eq:analytical_KLD}
	K(p \| q)= 
	\frac{1}{2} \left[ \Tr \left (\Sigma_q^{-1}\Sigma_{p}\right ) + (\mu_q - \mu_p)^\intercal \Sigma_q^{-1}(\mu_q - \mu_p) - n + \log \left(\frac{\det \Sigma_{q}}{\det \Sigma_{p}} \right) \right]
\end{equation}

Even though $K(p\|q)$ can be computed analytically if $p$ and $q$ are multivariate normally 
distributed, other probability distributions, such as Gaussian mixture models \cite{bishop2006pattern}, 
can be used to model the distribution of $\bar{r}$. Gaussian mixture models
are more flexible than multivariate normal distributions but have no analytical expression for the KL divergence. 
Even though there are methods to numerically approximate \eqref{eq:kl}, see for example \cite{hershey2007approximating}, 
it is a tradeoff between model fit and evaluation cost of $\mathcal{D}_{i,j}^*(p)$. 

For illustration, \Fig\ref{fig:batch_pdfs} shows an example of residual data from the case study. 
The upper plot shows that the normal distribution can approximate the mean and variation in 
batch data, except when there are outliers, and the variance is overestimated. One solution is to 
remove the extreme values, e.g., 10\% of the samples in each batch, by considering 
them as outliers before estimating the normal distribution. The lower plot in \Fig\ref{fig:batch_pdfs} 
shows the resulting normal distributions and it is visible that the estimated variance better captures 
the variation in data represented by a kernel density estimator. In this work, it is assumed that data 
in each batch can be represented by a multivariate normal distribution. 
\section{Approximated Distinguishability Measure Using Training Data}
\label{sec:modeling}
In many applications, the sets $\Omega_j$ are either partially, or completely, unknown because
training data only consist of a limited number of fault realizations. This means that 
the distinguishability measure \eqref{eq:dij_true} cannot be used. Instead, an 
approximated distinguishability measure is proposed where each fault mode 
is estimated from training data.  

\subsection{Distinguishability Measure for Data-Driven Analysis}

If training data are correctly labelled, partitioning the data into batches can be used to estimate 
pdfs belonging to different fault classes. The estimated pdfs belonging to 
fault class $f_j$ can be used to make a lower approximation 
of the true fault mode $\Omega_j$ denoted 
$\hat{\Omega}_j \subseteq \Omega_j$. Then, an approximation of 
\eqref{eq:dij_true} can be computed as
\begin{equation}
\mathcal{D}_{i,j}(p) = \min_{q \in \hat{\Omega}_j} K(p \| q)
\label{eq:dij}
\end{equation}
i.e., distinguishability is computed based on the set of already observed 
realizations of each fault $f_j$ available in training data. 

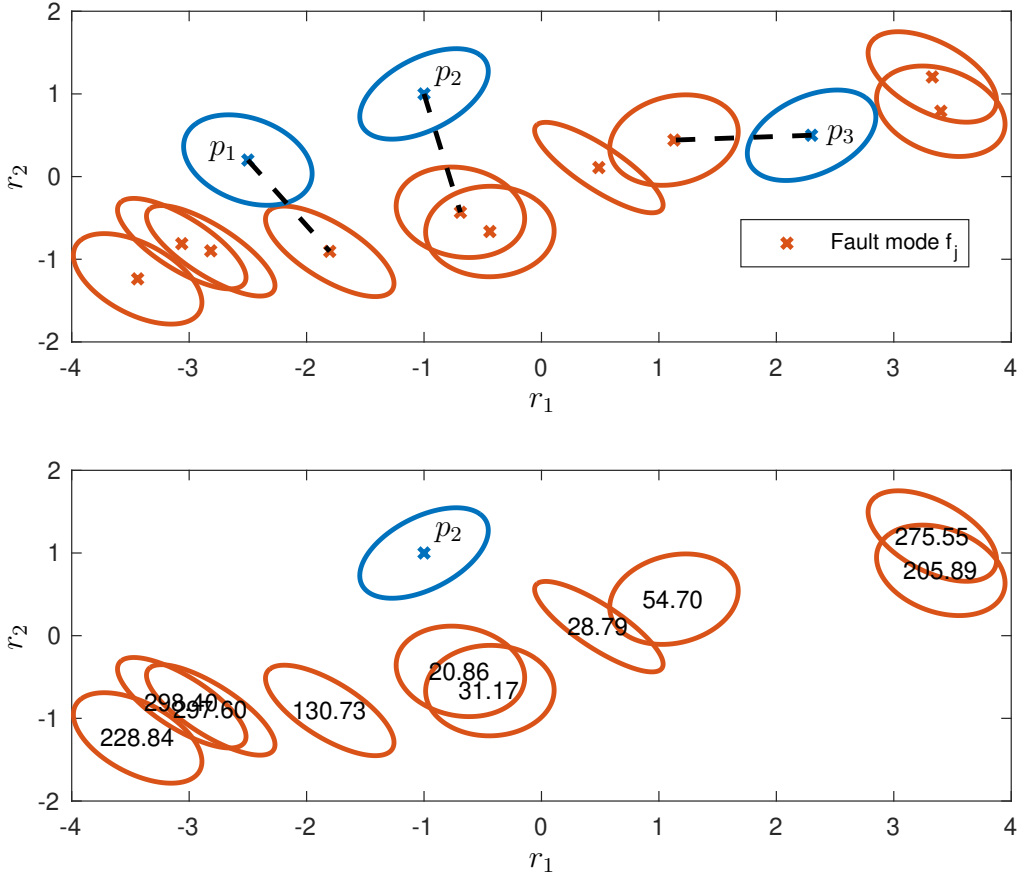
\begin{figure}[h!]
	\centering
	  \begin{tikzpicture}
    \node[anchor=south west,inner sep=0] (image) at (0,0) {
      \resizebox{0.95\columnwidth}{!}{\includegraphics{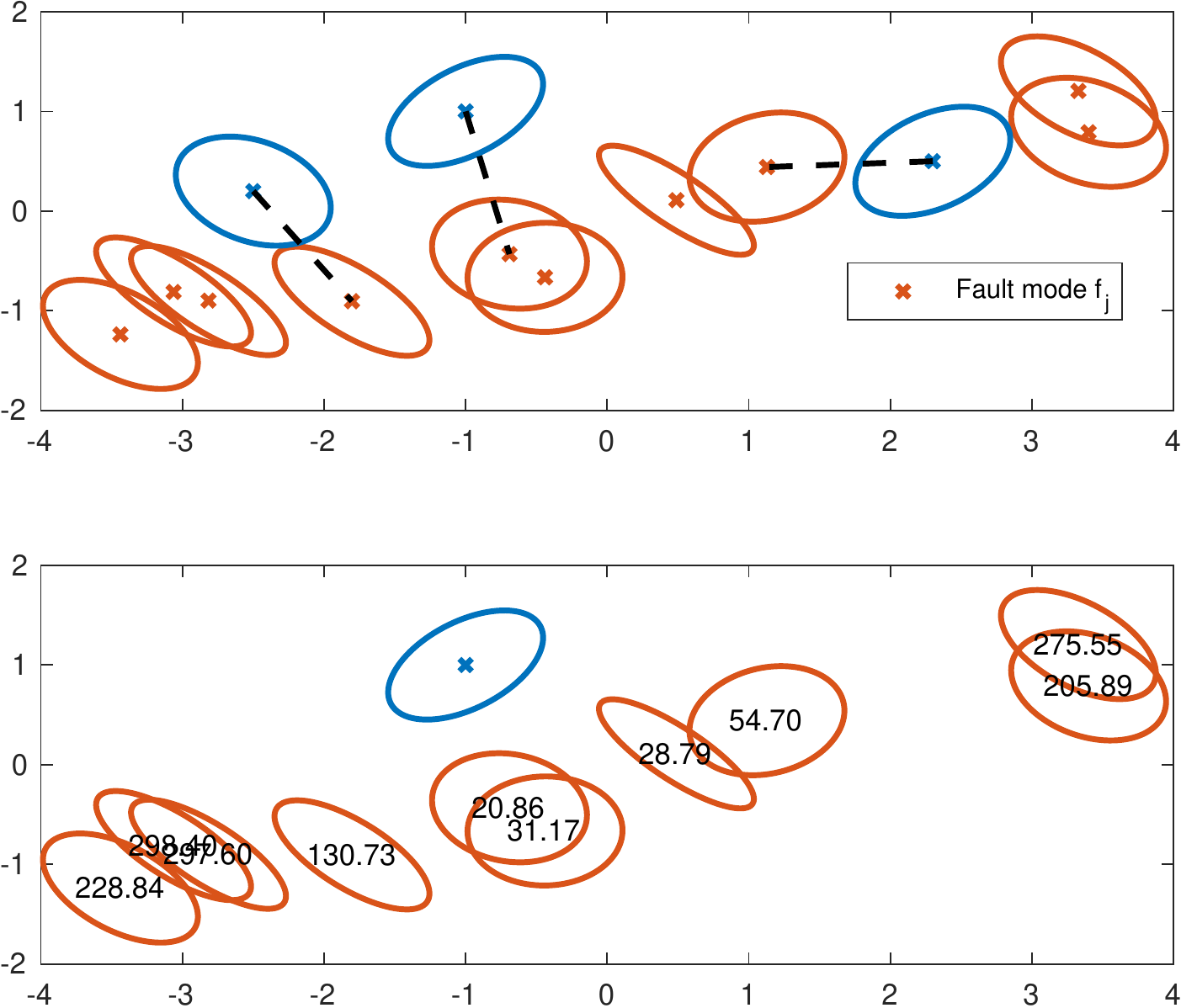}}
    };
    \begin{scope}[x={(image.south east)},y={(image.north west)}]
    \draw (0.19,0.82) node{ $p_1$};
     \draw (0.42,0.91) node{ $p_2$};
     \draw (0.82,0.84) node{ $p_3$};
     \draw (0.42,0.365) node{ $p_2$};
     \draw (-0.02,0.79) node{\rotatebox{90}{$r_2$}};
     \draw (0.515,0.52) node{$r_1$};
     \draw (-0.02,0.24) node{\rotatebox{90}{$r_2$}};
     \draw (0.515,-0.03) node{$r_1$};
     
    \end{scope}
  \end{tikzpicture}
	\caption{The upper plot shows an illustration of the approximate distinguishability measure 
	\eqref{eq:dij} from a set of pdfs, $p_1$, $p_2$, and $p_3$, to a fault mode $\hat{\Omega}_j$. 
	The dashed lines show each pdf $q$ that minimizes $\eqref{eq:dij}$ for each $p_i$. 
	For illustration, the computed KL divergence for each $q \in \hat{\Omega}_j$ is written 
	in the lower plot given $p_2$.}
	\label{fig:minkl}
\end{figure}

The approximate distinguishability measure \eqref{eq:dij} is illustrated in 
\Fig\ref{fig:minkl} where three pdfs, $p_1$, $p_2$, 
and $p_3$, are compared to a set of pdfs that is used to represent a fault mode $\hat{\Omega}_j$. The dashed lines show each 
pdf $q$ that minimizes $\eqref{eq:dij}$ for each $p_k$. The lower plot shows 
the computed KL divergence from one pdf, $p_2$, to all $q \in \hat{\Omega}_j$
which, in general, increases for pdfs $q$ that are located further away from $p_2$. 

Since $\hat{\Omega}_j \subseteq \Omega_j$, the relation between 
\eqref{eq:dij_true} and \eqref{eq:dij} is given by the inequality  
$0 \leq \mathcal{D}_{i,j}^*(p) \leq \mathcal{D}_{i,j}(p)$
which is derived from
\begin{equation}
\mathcal{D}_{i,j}^*(p) = \min_{q \in \Omega_j} K(p \| q) \leq \min_{q \in  \hat{\Omega}_j} K(p \| q) = \mathcal{D}_{i,j}(p)
\end{equation}
The approximation \eqref{eq:dij} gives an upper bound of how easy it is to 
reject $f_j$ for $f_i$ given $p$. Note that \eqref{eq:dij_det_isol} also holds for \eqref{eq:dij}, i.e.,
$\mathcal{D}_{i,NF}(p) \geq \mathcal{D}_{i,j}(p)$.

\section{Open-Set Fault Classification Using Distinguishability}
\label{sec:classification}

Since different faults can have similar impact on system operation, it is 
relevant to not only select the most likely fault class but to identify all fault 
classes that can explain a set of observations. Here, a set of $m$ one-class 
classifiers is used to model data from each of the $m$ fault classes to see if 
each class can explain the observation or not. If a pdf $p$ cannot be explained 
by any of the known fault classes, i.e., $p \not\in \hat{\Omega}_j$ for all $f_j$, 
it belongs to an unknown fault class. Note that there can be two types of unknown 
faults \cite{scheirer2014probability}: 
\begin{itemize}
\item Data belong to a new unknown fault class, or 
\item Data is a new realization of a known fault class $f_j$, i.e., $p \in \Omega_j \setminus \hat{\Omega}_j$.   
\end{itemize}
These unknown fault scenarios require extra attention, for example, by a technician, to identify 
the root cause and correctly label data to be used for updating the corresponding fault mode. 

\subsection{One-class Classification}

Here, a one-class classifier is proposed using the approximate distinguishability measure \eqref{eq:dij} 
to tests if a new pdf $p$ can be explained by a fault mode $f_j$ or not. The proposed classifier is 
denoted $\mathcal{D}_j(p) =  \min_{q \in \hat{\Omega}_j} K(p \| q)$ with respect to \eqref{eq:dij} to 
emphasize that the class label of $p$ is unknown. Hereafter, the classifier modeling a fault mode $f_j$ 
will be referred to in the text as $\mathcal{D}_j$. 

An open set classification algorithm for fault diagnosis is then formulated using one $\mathcal{D}_j$ 
classifier for each known fault class $f_j$. Fault classification is then performed such that the fault 
hypothesis $f_j$ is rejected if  
\begin{equation}
\mathcal{D}_j(p) > J_j
\label{eq:Dj_J}
\end{equation}
where $J_j$ is a threshold. A small threshold $J_j$ increases the risk of falsely rejecting the true fault 
class while a large threshold $J_j$ means that fault $f_j$ is more likely to be a wrong hypothesis 
increasing fault classification ambiguity.

\subsection{Tuning of The One-Class Classifier Thresholds Using Within-Class Distinguishability}

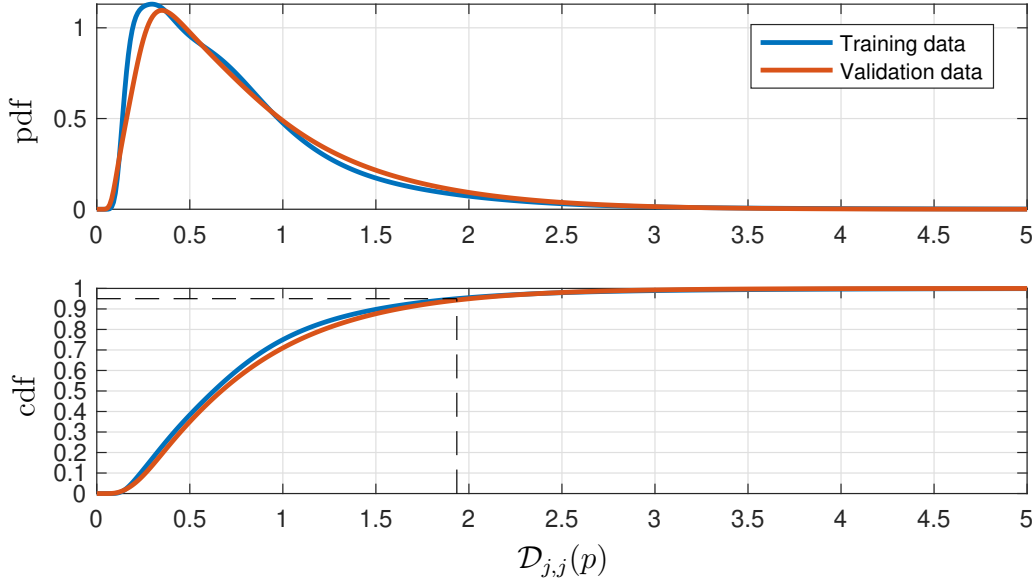
\begin{figure}[h!]
	\centering
	  \begin{tikzpicture}
    \node[anchor=south west,inner sep=0] (image) at (0,0) {
      \resizebox{0.95\columnwidth}{!}{\includegraphics{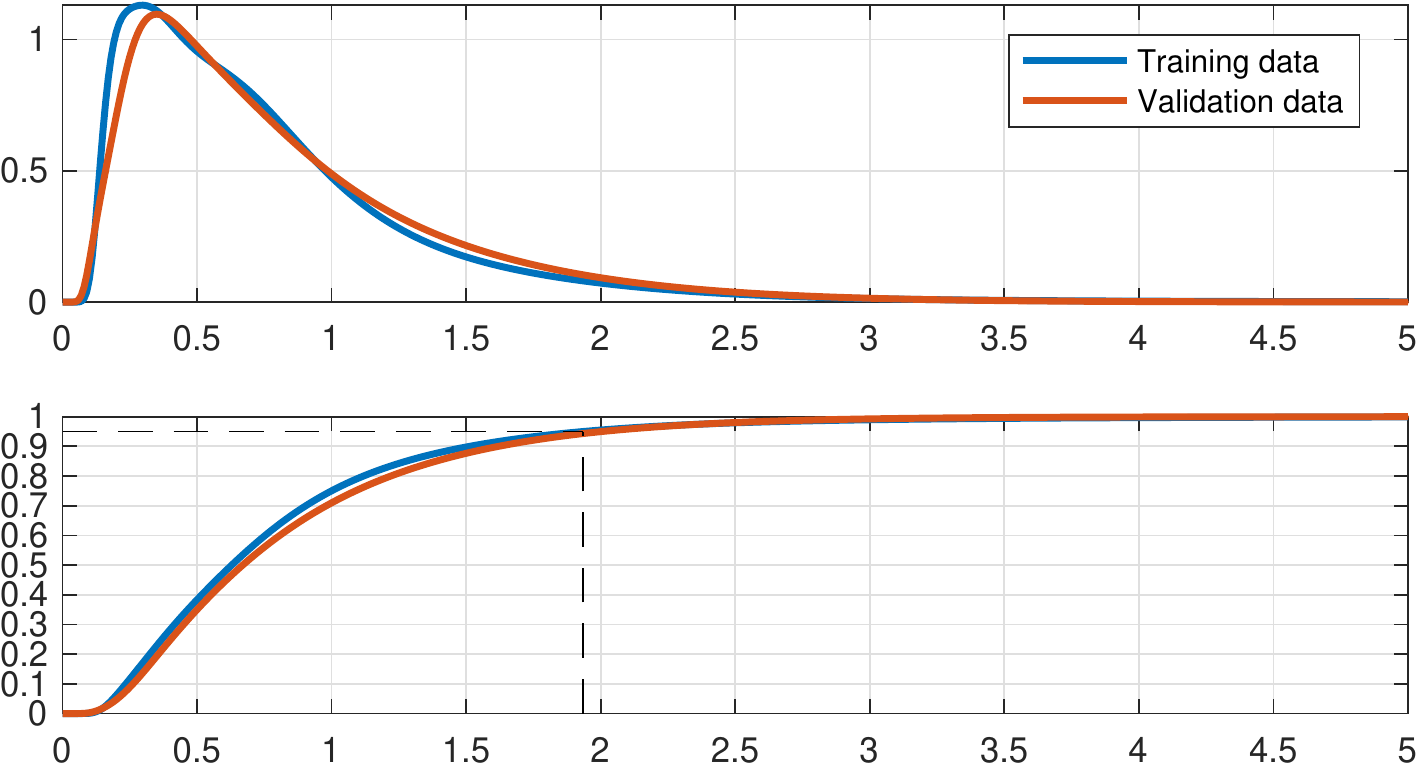}}
    };
    \begin{scope}[x={(image.south east)},y={(image.north west)}]
    \draw (0.52,-0.06) node{$\mathcal{D}_{j,j}(p)$};
     \draw (-0.03, 0.77) node{\rotatebox{90}{pdf}};
     \draw (-0.03, 0.25) node{\rotatebox{90}{cdf}};
    \end{scope}
  \end{tikzpicture}
	\caption{Kernel density estimation of within-class distinguishability for the fault-free class (pdf in upper plot and cdf in lower plot). Dashed line represents threshold $J_{NF}$ tuned to have a $5\%$ outlier rate.}
	\label{fig:ksdensity_J}	
\end{figure}

A tuning strategy of the threshold $J_j$ in \eqref{eq:Dj_J} is proposed such that most of the 
pdf's $p \in \hat{\Omega}_j$ should be explained by fault class $f_j$ if $p$ is removed 
from $\hat{\Omega}_j$. Let 
\begin{equation}
\mathcal{D}_{j,j}(p) = \min_{q \in (\hat{\Omega}_j \setminus \{p\})} K(p \| q)
\label{eq:within_class_dij}
\end{equation} 
which is here referred to as \emph{within-class distinguishability}. Analyzing the 
distribution of $\mathcal{D}_{j,j}(p)$ for all $p \in \hat{\Omega}_j$ can be used 
to select a threshold. The distribution will have non-negative support and is here 
approximated using a kernel density estimation method \cite{hastie2009elements} as illustrated 
in \Fig~\ref{fig:ksdensity_J}. Note that \eqref{eq:within_class_dij} does not state anything 
about the relation between the sets $\hat{\Omega}_j$ and $\Omega_j$ but 
rather gives information about how scattered training data are from that fault class. 
Let $\Phi(x)$ denote the cumulative density 
function (cdf) of the estimated distribution and let $\alpha$ denote a desired 
false alarm rate. Then, the threshold $J_j$ is selected such that 
$\Phi(J_j) = 1-\alpha$. The lower plot in \Fig~\ref{fig:ksdensity_J} illustrates 
selecting a threshold corresponding to $\alpha = 5\%$.

\subsection{Computing Fault Hypotheses}
\label{sec:computing_fault_hypotheses}
If $\mathcal{D}_j(p)$ is large, it means that $p$ is not likely to be explained by fault $f_j$.
If $\mathcal{D}_j(p)$ is large for all fault classes $f_j$, this means that no known fault class can explain $p$, 
thus indicating the occurrence of an unknown fault class. 
These results give a systematic approach to compute fault hypotheses, including 
the unknown fault case, by evaluating and comparing $\mathcal{D}_j(p)$ for each known fault class 
$f_j$. 

The classifier (diagnosis) output $D$ given an observation $p$ is determined using the following decision rule: 
\begin{enumerate}
\item If $\mathcal{D}_{NF}(p) \leq J_{NF}$ then $D = \{NF\}$, i.e., the output of the classifier is that the system is fault-free. 
\item If $\mathcal{D}_{NF}(p) > J_{NF}$ then $D = \{f_j : \mathcal{D}_{j}(p) < J_{j}\}$, i.e., the output is the set of all known 
fault classes that can explain $p$. 
\item If $\mathcal{D}_{j}(p) > J_{j}$ for all known fault classes $f_j$, i.e., no known fault class can explain $p$, then 
$D = \{f_x\}$ where $f_x$ denotes an unknown fault class. 
\end{enumerate}

An advantage of testing the fault hypothesis for each fault class individually, is that there is no bias in 
the diagnosis output if the fault models are trained using imbalanced datasets. A fault $f_j$ is a diagnosis candidate if $\mathcal{D}_j(p)$ 
does not exceed the threshold $J_j$. Note that when the output $D$ contains at least one known fault class, 
it is still plausible that an unknown fault type has occurred, thus $f_x$ is always a plausible fault hypothesis. 
However, $f_x$ is only explicitly stated as a diagnosis output if $\mathcal{D}_j(p)$ is large for all $f_j$. 
If the $NF$ class is not rejected, i.e., if $\mathcal{D}_{NF}(p) \leq J_{NF}$, there could still
be an undetected fault present in the system since $\hat{\Omega}_{NF} \subseteq \hat{\Omega}_j$ 
for all $f_j$. However, if the $NF$ class is not rejected, it is more likely that the system is fault-free 
than faulty. Therefore, the only diagnosis output in this case is that the system is fault-free. 

\section{Data-driven Fault Size Estimation}
\label{sec:estimation}

The method presented in Section~\ref{sec:classification} provides 
a means to classify new data, but it does not give any information 
about the severity of these faults. If each pdf $q_k \in \hat{\Omega}_j$  
has a known fault size $\theta_k$, this information can be utilized to 
estimate the size of new realizations of fault $f_j$ by comparing how 
similar the distribution of new data is to training data. One approach that has 
been suggested in 
\cite{Grezmak2019} is to model faults into qualitative classes, such as 
\{normal, slight, large\}. Another way, which is a method that is largely 
unexplored, is to find a quantitative severity estimation $\hat{\theta}$. 
Here, it is assumed that two pdfs $p$ and $q$ are from the same fault class 
$f_i$ and have similar fault sizes $\theta_p \simeq \theta_q$ should also be 
similar in a KL divergence sense, i.e., $K(p\|q)$ should be small. Fault size 
estimation is formulated as a convex optimization problem by using the 
KL divergence as a dissimilarity measure. 

The fault size of a pdf $p$ is estimated by finding a representation 
of $ p $ using a set of training distributions. Assume that for a given $p$, 
$f_j \in D$ is a fault hypothesis. Then, the fault size $\theta$ of the corresponding fault 
class $ f_j $ is estimated from a linear combination of pdfs 
$\{q_1,q_2,\dots,q_M\} =  \hat{\Omega}_j$, where $M = |\hat{\Omega}_j|$, 
by solving the following optimization problem:
\begin{equation}
\begin{aligned}
	\lambda^*_{1},...,\lambda^*_{M} = \arg\min_{\lambda_{1},...,\lambda_{M}} &&& K(p\|\lambda_{1}q_{1} + ... + \lambda_{M}q_{M})\\
	\text{s.t.} &&& \sum_{k = 1}^{M} \lambda_{k} = 1 \\
	&&& \lambda_{k}\geq 0, \quad \forall k = 1,2, \ldots, M
\end{aligned}\label{eq:lambda_est} 
\end{equation}
The fault size estimate $ \hat{\theta} $ is then computed as a weighted sum of the 
fault sizes $ \theta_1, \theta_2, \ldots, \theta_M $, corresponding to the pdfs 
$ q_1, q_2,  \ldots q_M $, as $\hat{\theta} = \sum_{k=1}^M \lambda^*_k \theta_k$.
The estimate $ \hat{p} $ in \eqref{eq:lambda_est} is obtained by using all pdfs 
$q_k \in \hat{\Omega}_j $ in the optimization. This is an ineffective strategy 
since it increases the computational cost by adding numerous distributions $q_k$ 
likely to correspond to $\lambda_k = 0$. If multiple datasets have been collected from a variety of 
severities and conditions, it is unlikely that batches of new data would have a distribution that 
is similar to all pdfs in the training set. Using this line of reasoning, only a 
small subset of all pdfs is reasonably of interest when formulating the optimization problem \eqref{eq:lambda_est}. 

Let $\{q_{(1)}, q_{(2)}, \ldots, q_{(M)}\} = \hat{\Omega}_j$ denote an ordered set of all elements  
such that $K(p\|q_{(1)}) \leq K(p\|q_{(2)}) \leq \ldots \leq K(p\|q_{(M)})$. Then $\hat{\Omega}_j^l \subseteq \hat{\Omega}_j$
is defined as the first $l$ elements in the ordered set to reduce the number of parameters in \eqref{eq:lambda_est}.
The parameter $ l \leq M $ is the cardinality of $ \hat{\Omega}_j^l $ and is calibrated 
to include the subset of elements in $\hat{\Omega}_j^l$ that are "similar" to $p$, i.e., $K(p\|q)$ is 
relatively small and similar in value to $K(p\|q_{(1)})$.

If $q_{(1)}, q_{(2)}, \ldots, q_{(l)}$ are multivariate normal distributions then 
$\sum_{k=1}^{l} \lambda_k q_{(k)}$ is a Gaussian mixture model \cite{hastie2009elements}.
Here, Monte Carlo sampling is used to estimate $K(p\| q )$, where $q = \sum_{k=1}^{l} \lambda_k q_{(k)}$, as
\begin{equation} \label{eq:kld_mc}
K_{MC}(p || q) = \frac{1}{v} \sum_{\gamma = 1}^{v} \log\left( \frac{p(x_\gamma)}{q(x_\gamma)} \right)
\end{equation} 
by generating $v$ samples $\{ x_\gamma \}_{\gamma = 1}^{v}$ 
from $p$ to approximate the integration in~\eqref{eq:kl}.
By the law of large numbers $\lim_{v \to \infty} K_{MC}(p||q) = K(p||q)$. 
A closer examination of the actual upper and lower bounds of this approximation is found in 
\cite{durrieu2012lower}.

Applying \eqref{eq:kld_mc} in \eqref{eq:lambda_est} gives the updated algorithm:
\begin{equation}
\begin{aligned}
	\lambda^*_{1},...,\lambda^*_{l} = &&& \\
	\arg\min_{\lambda_{1},...,\lambda_{l}} &&& \frac{1}{v} \sum_{\gamma = 1}^{v} \log \left( \frac{p(x_\gamma)}{\lambda_{1}q_{(1)}(x_\gamma) + ... + \lambda_{l}q_{(l)}(x_\gamma)} \right)\\
	\text{s.t.} &&& \sum_{k = 1}^{l} \lambda_{k} = 1 \quad \lambda_{k}\geq 0, \quad \forall k = 1,2, \ldots, l
\end{aligned}
\label{eq:KLD_minimization}
\end{equation}
where $\{q_{(1)}, q_{(2)}, \ldots, q_{(l)}\} = \Omega_j^l$ and the fault size is estimated as $\hat{\theta} = \sum_{k=1}^l \lambda^*_k \theta_k$. Note that if the diagnosis output $D$ contains multiple fault hypotheses $f_j$, a fault size is estimated for each fault hypothesis by solving \eqref{eq:KLD_minimization} for each $f_j \in D$.
\section{Case study}
\label{sec:casestudy}
\begin{figure}[h!]
	\centering
	\includegraphics[width=0.9\linewidth]{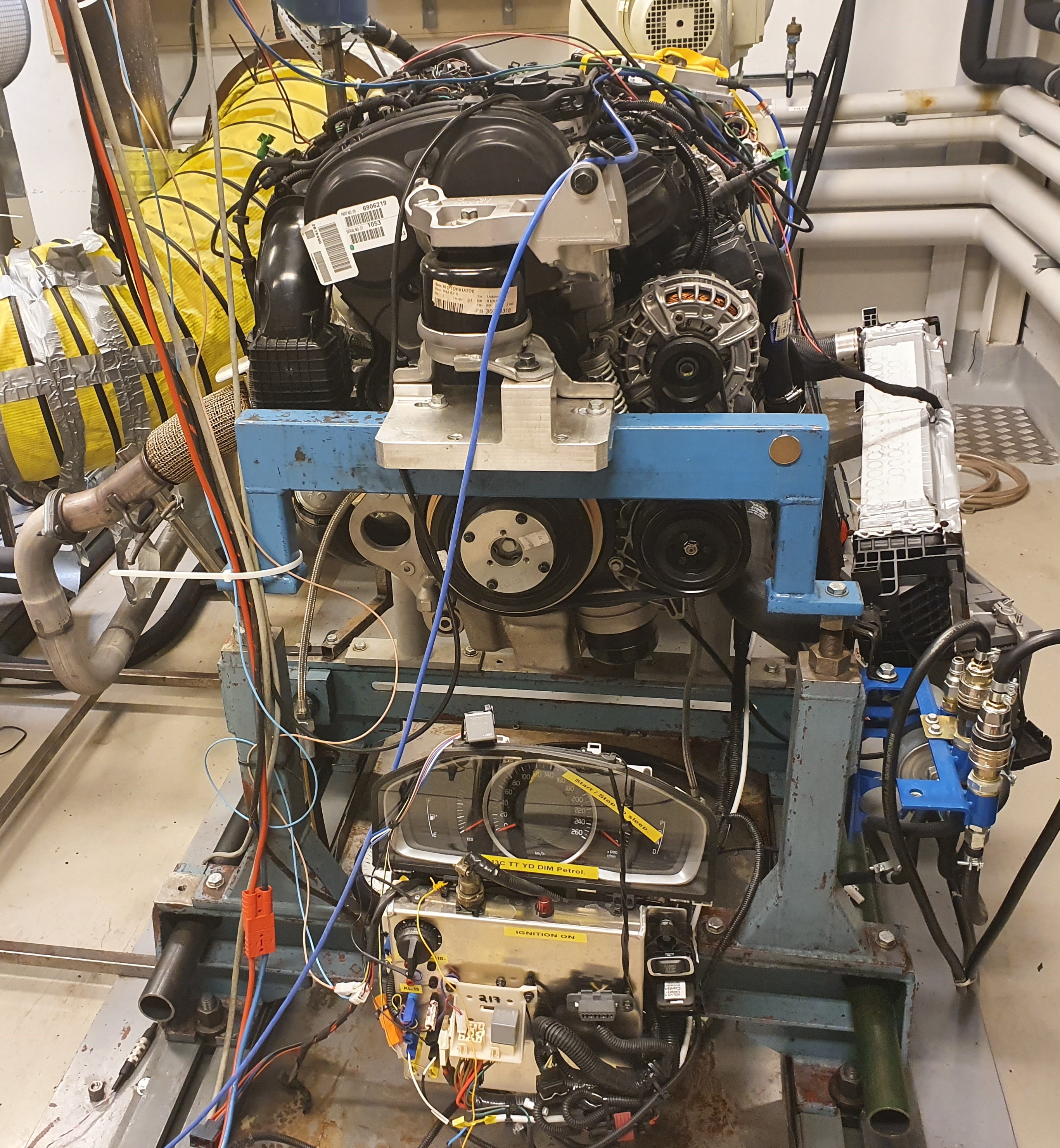}
	\caption{The engine test bench which was used for data collection. The engine is a commercial four-cylinder combustion engine with standard sensor and actuator configuration \citep{jung2020residual}.}
	\label{fig:engine_test_bench}
\end{figure}

\begin{figure}[h!]
	\centering
	\begin{tikzpicture}
	\node[anchor=south west,inner sep=0] (image) at (0,0) {
		\resizebox{0.9\linewidth}{!}{\includegraphics{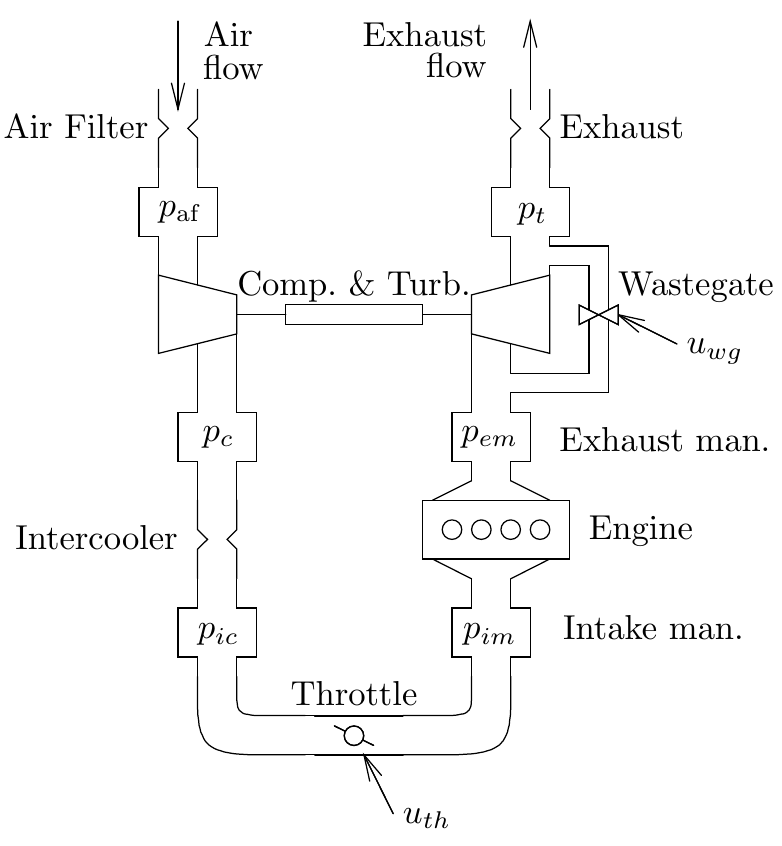}}};
	\begin{scope}[x={(image.south east)},y={(image.north west)}]
	\fill[white!0!white] (0.4,0) rectangle (0.6,0.105);
	\fill[white!0!white] (0.7917,0.5) rectangle (1.0,0.64);
	\draw[->] (0.22, 0.22) -- (0.17, 0.21) node[left]{$y_{pic}$};
	\draw[->] (0.22, 0.26) -- (0.17, 0.27) node[left]{$y_{Tic}$};
	\draw[->] (0.68, 0.21) -- (0.73, 0.20) node[right]{$y_{pim}$};
	\draw[->] (0.19, 0.80) -- (0.12, 0.79) node[below]{$y_{Waf}$};
	\draw[->] (0.53, 0.36) -- (0.48, 0.35) node[left]{$y_{\omega}$};
	\draw[->] (0.46, 0.10) -- (0.47, 0.06) node[right]{$y_{xpos}$};
	\draw[] (0.00, 0.58) node[right]{$y_{pamb}$};
	\draw[] (0.00, 0.52) node[right]{$y_{Tamb}$};
	\draw[<-] (0.7917,0.62) -- (0.8417,0.61) node[right]{$u_{wg}$};
	\draw[<-] (0.53, 0.39) -- (0.48, 0.41) node[left]{$u_{mf}$};
	\end{scope}
	\end{tikzpicture}
	\caption{A schematic of the model of the air flow through the model. Available output signals are sensors $y$ and actuators $u$. The figure is used with permission from \cite{Eriksson2002}.}
	\label{fig:engine_schematic}
\end{figure}
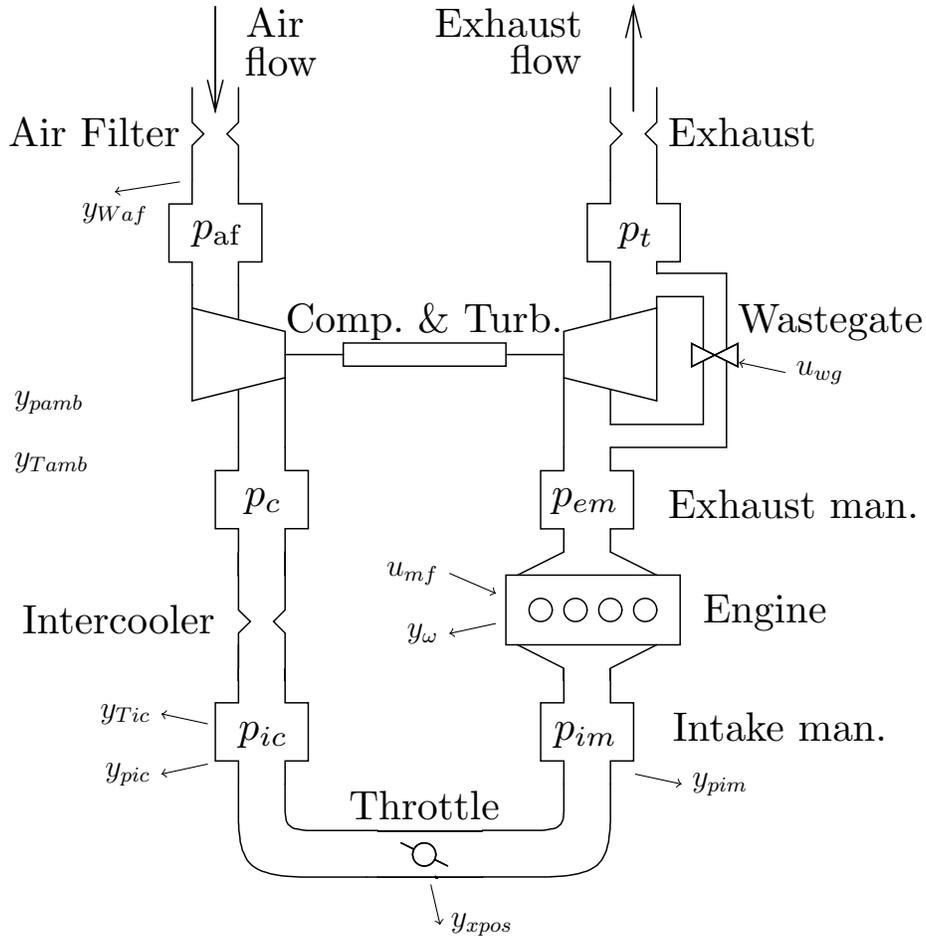

The diagnostic framework is evaluated by using experimental data collected from
an engine test bench, see \Fig\ref{fig:engine_test_bench}. The engine is a commercial, turbo charged, four-cylinder,
internal combustion engine from Volvo Cars. The sensor and actuator setup are the standard 
commercial configuration for the engine \cite{jung2018combining}. 
Figure~\ref{fig:engine_schematic}  shows a schematic view of the engine along 
with the monitored signals where $y$ denotes sensor measurements and $u$ 
denotes actuator signals. The system represents the air path through the engine
and is an interesting case study for fault 
diagnosis because of its non-linear dynamic behavior and wide operating range. 
In addition, the coupling from the exhaust flow to the air intake by the turbocharger 
complicates fault isolation since the effect of a fault anywhere in the system will 
affect the behavior of many other components in the whole system. 

\subsection{Data Collection}
Engine sensor data are collected from various operating scenarios including different types of faults and fault magnitudes. 
The fault classes include four multiplicative sensor faults, a leakage in the intake manifold after the throttle, as well as nominal 
system operation, see Table~\ref{tab:fault_classes}. 
The sensor faults are introduced by altering the sensor output gain in the engine control system. 
Since the errors are injected in this way, the faulty signal output is used in the engine control 
scheme which gives a more realistic fault realization compared to if the error is simulated in the data 
using post processing. Each sensor fault is injected by multiplying the measured variable $x_i$ in 
each sensor $y$ by a factor $\theta$ such that the resulting output is given as 
$y = (1 + \theta) x$ where $\theta$ is the fault size and $\theta = 0$ corresponds to the 
nominal case. The leakage in the intake manifold is introduced by opening valves with different diameters 
during operation.

\begin{table}[h!]
	\caption{Fault classes considered in the case study. All sensor faults are induced as multiplicative faults.}
	\label{tab:fault_classes}
	\centering
	\begin{tabular}{cl}
		\hline
		Fault Class & Description\\
		\hline
		$ NF $  & Fault-free class \\
		$ f_{ypim} $ & Fault in intake manifold pressure sensor \\
		$ f_{ypic} $ & Fault in intercooler pressure sensor \\
		$ f_{ywaf} $ & Fault in air-mass flow sensor \\
		$ f_{iml} $ & Leakage in the intake manifold \\
		\hline
	\end{tabular}
\end{table}
 
\begin{figure}[h!]
	\centering
	\includegraphics[width=\linewidth]{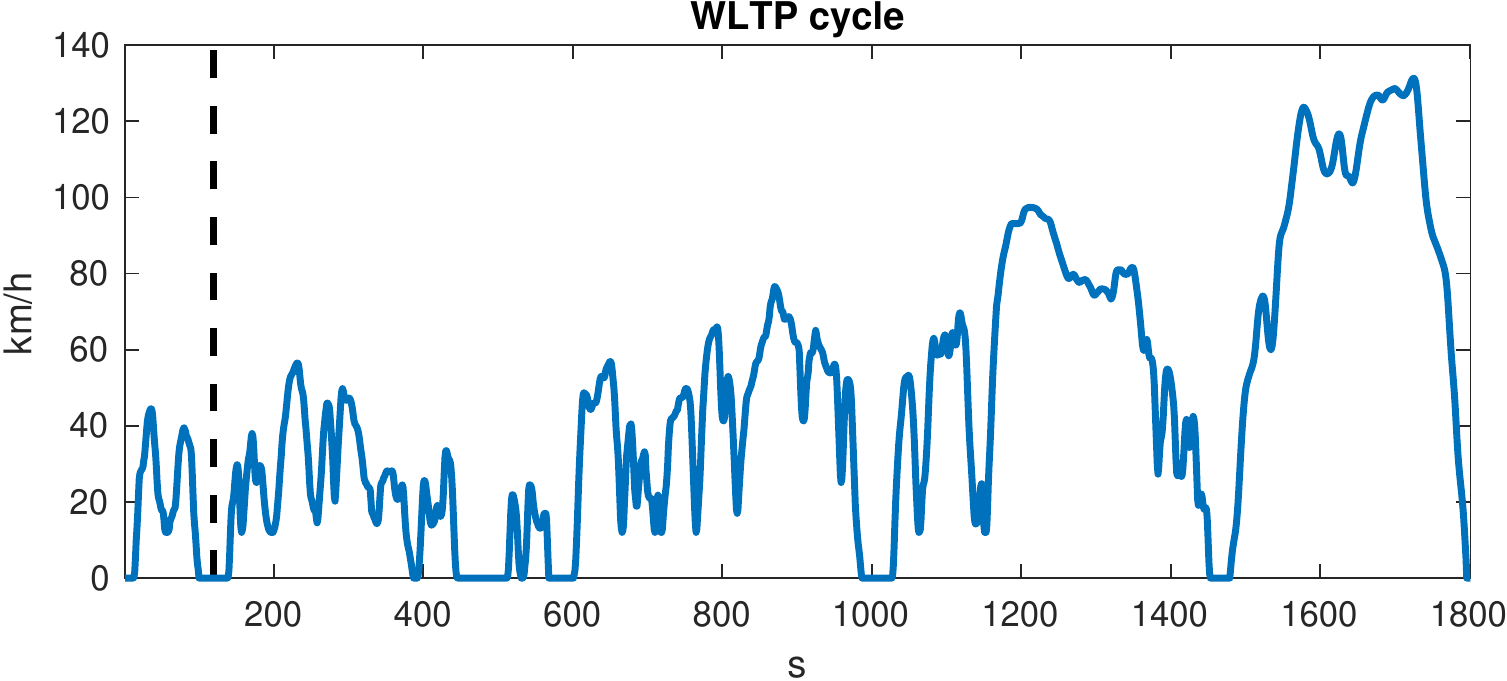}
	\caption{Speed profile for the WLTC class 3 test cycle. The dashed line represents the time when a fault is introduced in each fault scenario.}
	\label{fig:WLTC}
\end{figure}

Each dataset was collected during transient operation following the class~3 Worldwide harmonized Light-duty 
vehicles Test Cycle (\textsc{WLTC}), which is part of the World harmonized 
Light-duty vehicles Test Procedure (\textsc{WLTP}) \cite{Tutuianu2013}. 
The cycle is shown in \Fig\ref{fig:WLTC} and is used since it covers a variety of operating 
conditions. One dataset has been collected for each fault class and fault size in Table~\ref{tab:fault_modes} 
resulting in 26 datasets (24 fault scenarios and two fault-free datasets). Each fault is introduced in 
the dataset after approximately two minutes of the driving cycle. 

\begin{table}[h!]
	\caption{Fault classes and known magnitudes (i.e., $\theta$ for multiplicative sensor faults and leakage diameters) represented in training data. Data from the leakage $f_{iml}$ have been collected from two known diameters of the orifice.}
	\label{tab:fault_modes}
	\centering
	\begin{tabular}{ccccccccc}
		\hline
		Fault Class &  \multicolumn{8}{c}{Fault magnitudes} \\
		\hline 
		 $ NF $ & \\
		 $ f_{ypim} $ &  -20\%& -15\% & -10\% & -5\% & 5\% & 10\% & 15\% & \\
		 $ f_{ypic} $ & -20\% & -15\% & -10\% & -5\% & 5\% & 10\% & 15\% &\\
		 $ f_{ywaf} $ & -20\% & -15\% & -10\% & -5\% & 5\% & 10\% & 15\% & 20\%\\
		 $ f_{iml} $ & 4mm &  6mm\\
		\hline
	\end{tabular}
\end{table}

\subsection{Residual Generation}

The proposed method can be applied to any set of features to be used for 
fault diagnosis. In dynamic systems operated in various transient operating 
conditions, such as the engine, using raw sensor data as features requires that a data-driven 
classifier captures these dynamics since these signals can vary significantly over time. 
Here, a set of four residual generators $\bar{r} = (r_1, r_2, r_3, r_4)$ is generated 
by comparing predictions from a set of Recurrent Neural Networks (RNN) with the corresponding sensor 
outputs, see \Fig\ref{fig:residual}, that will be used as features for fault diagnosis in the case study. A 
summary of the set of residual generators used in the case study is 
presented here. For the interested reader, a more detailed description is given in \cite{jung2020residual}.

\tikzstyle{block} = [rectangle, draw, fill=blue!10, 
    text width=4em, text centered, minimum height=3.0em]
\tikzstyle{wide_block} = [rectangle, draw, fill=blue!10, 
    text width=6.0em, text centered, minimum height=2.5em]
\tikzstyle{line} = [draw, -latex']
\tikzstyle{sum} = [draw, fill=blue!10, circle, node distance=1cm]
\begin{figure}[h!]
\centering
  \begin{tikzpicture}[node distance=0.2cm and 0.2cm]
	\node [wide_block] (system) {\includegraphics[width=55pt]{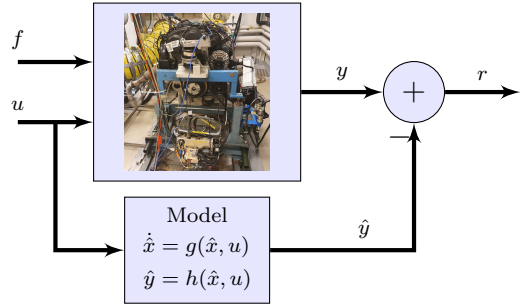}};
	\node [block, below= of system] (model) {\scriptsize Model \\ \vspace{0.1cm} $\begin{aligned} \dot{\hat{x}} &= g(\hat{x}, u) \\ \hat{y} &= h(\hat{x},u) \end{aligned}$};
	\draw ($(system.east) + (1.5,0.0)$) node[sum] (sum) {+};
	\draw[line, ultra thick]  ($(system.west) + (-1.0,0.40)$) node[above]{\scriptsize $f$} -- ($(system.west) + (-0.4,0.40)$) --  ($(system.west) + (0.0,0.40)$);
	\draw[line, ultra thick]  ($(system.west) + (-1.0,-0.40)$) node[above]{\scriptsize $u$} -- ($(system.west) + (-0.75,-0.40)$) --  ($(system.west) + (0.0,-0.40)$);
	\draw[line, ultra thick]  ($(system.west) + (-0.5,-0.40)$) |-  ($(model.west)$);
	\draw[line, ultra thick] (system)  -- node[above]{\scriptsize $y$} (sum);
	\draw[line, ultra thick] (model)  -- ($(model.east) + (0.75,0.0)$) -- ($(model.east) + (1.25,0.0)$) node[above]{\scriptsize $\hat{y}$} -| (sum);
	\draw[line, ultra thick] (sum)  -- node[above]{\scriptsize $r$} ($(sum.east) + (1.00,0.0)$);
	
	\draw[] ($(sum.east) + (-0.60,-0.60)$) node{$-$};
	  \end{tikzpicture}
  \caption{An example of a residual $r$ comparing measurements from the system $y(t)$ with model predictions $\hat{y}$.}
  \label{fig:residual}
\end{figure}  

\begin{figure}[h!]
	\centering
	\includegraphics[width=\linewidth]{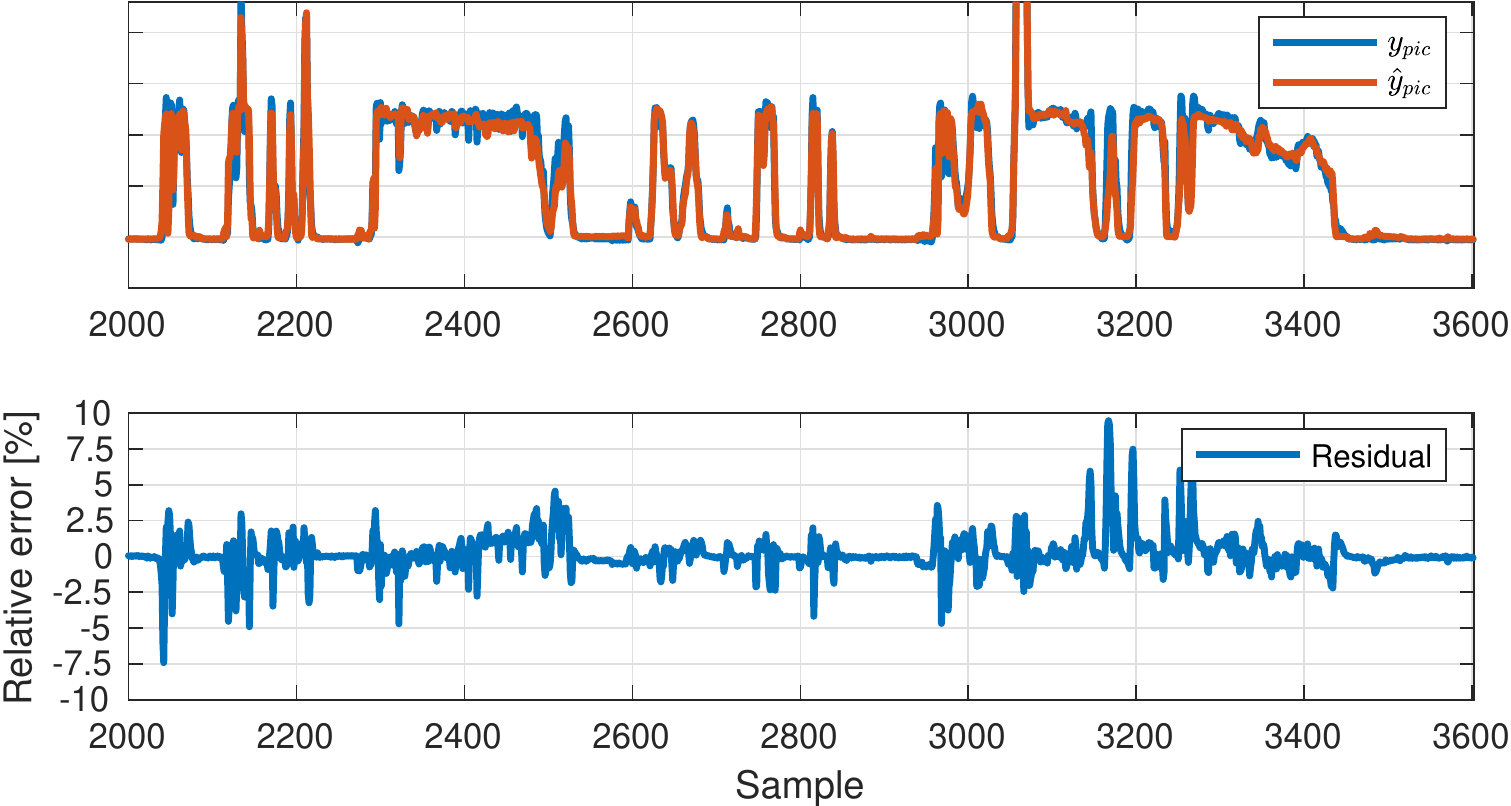}
	\caption{The upper plot compares data from sensor $y_{pic}$ and model predictions from an RNN regression model. The lower plot shows the resulting residual $r_3$.}
	\label{fig:r3_nf}
\end{figure}

\begin{figure}[h!]
	\centering
	\includegraphics[width=\linewidth]{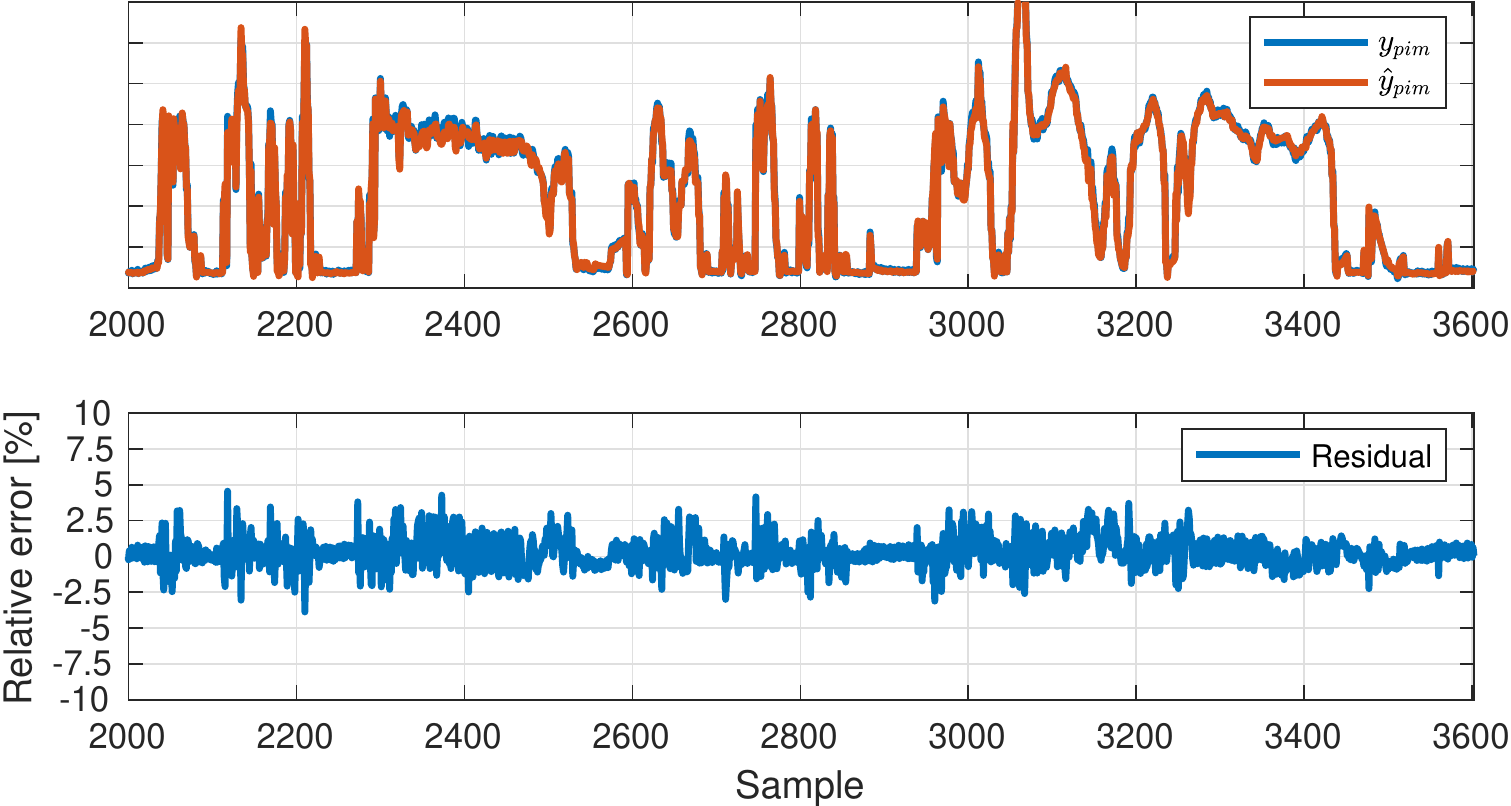}
	\caption{The upper plot compares data from sensor $y_{pim}$ and model predictions from an RNN regression model. The lower plot shows the resulting residual $r_4$.}
	\label{fig:r4_nf}
\end{figure}

The prediction performance of two of the four residual generators is shown 
in \Fig\ref{fig:r3_nf} and \Fig\ref{fig:r4_nf}, respectively. The figures show that the residual 
generators filter out most of the system dynamics and have a small relative 
prediction error. To show the impact of different faults on the residual output,
three of the four residuals are plotted against each other for different fault 
classes in \Fig\ref{fig:residual_faults}. The different faults are 
projected into different directions in the residual space which indicates that it is 
possible to distinguish between these faults. However, some fault 
classes are partially overlapping, e.g., a fault in the sensor measuring pressure 
after the intake manifold, $f_{ypim}$, and a leakage in the intake manifold, 
$f_{iml}$. It is expected that it is more difficult to distinguish between these 
two faults since they are related to the pressure after the throttle.  

\section{Evaluation}
\label{sec:evaluation}

The proposed methods for quantitative fault diagnosis analysis and 
open-set fault classification are evaluated using data from the engine 
case study. Residual data from all fault scenarios in Table~\ref{tab:fault_modes} 
are partitioned into batches where each batch is used to estimate a multivariate 
normal distribution. Different batch sizes are tested to evaluate the effect on 
classification performance.
First, the distinguishability measure \eqref{eq:dij} is used to evaluate fault 
detection and isolation performance. Then, the proposed $\mathcal{D}_j$ 
classifier is evaluated, including classification of unknown faults and fault size 
estimation.
   
\subsection{Data Processing}

To analyze how the batch size will impact fault diagnosis performance, outputs from the 
four residual generators are partitioned into batches of various lengths in the interval 
of 50 - 300 samples. For each interval, the mean and covariance 
matrix of a four-dimensional multivariate normal distribution are estimated. 

\begin{figure}
\centering
  \begin{tikzpicture}
    \node[anchor=south west,inner sep=0] (image) at (0,0) {
      \resizebox{0.93\columnwidth}{!}{\includegraphics{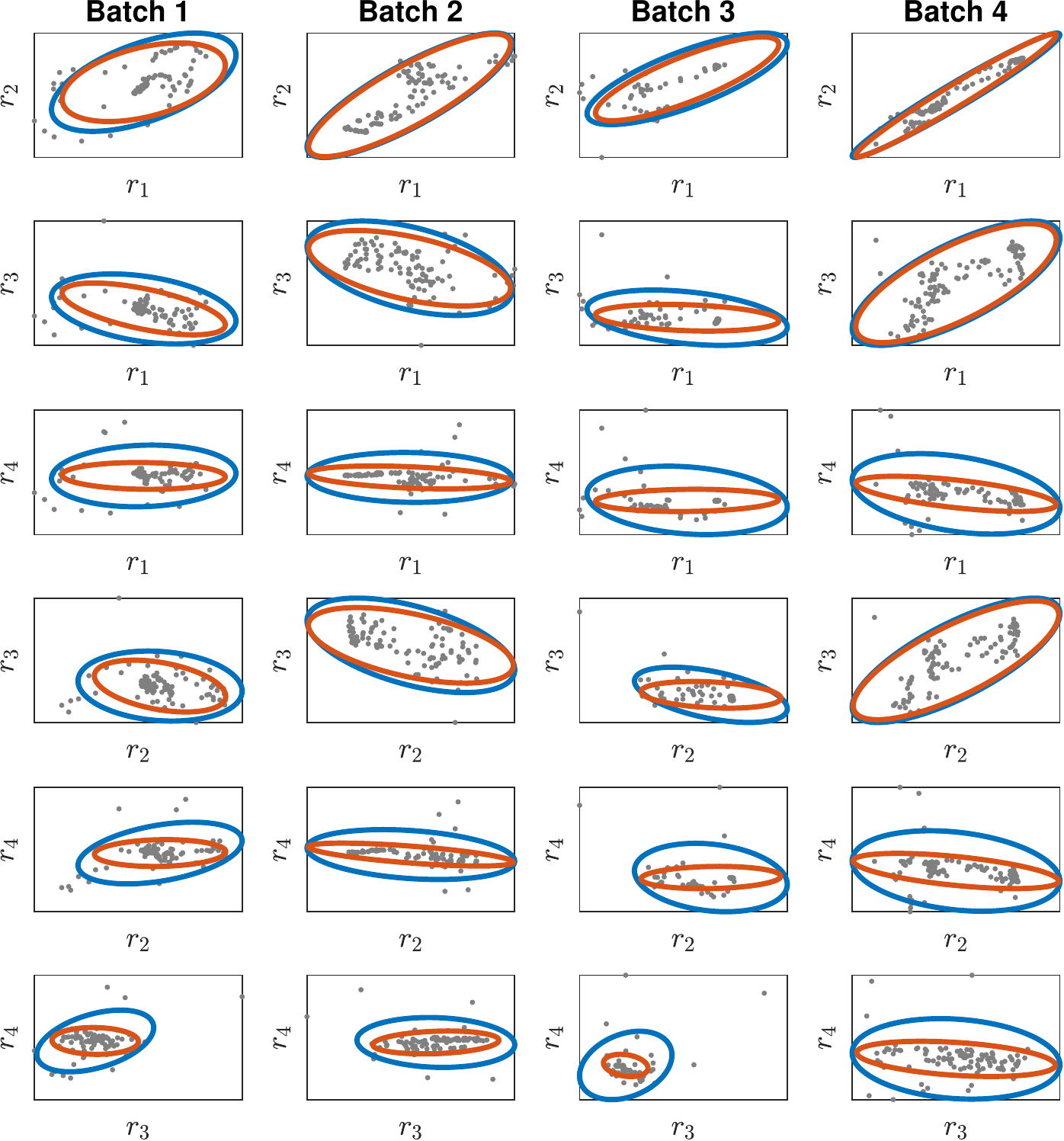}}
    };
    \begin{scope}[x={(image.south east)},y={(image.north west)}]
    \end{scope}
  \end{tikzpicture}
	\caption{Residual data from four different batches compared to estimated multivariate normal distributions. The blue ellipses
	correspond to the covariance estimated using the whole batch data. The orange ellipses are estimated using a subset of 
	batch data after removing 10\% of the samples that are outliers.} 
	\label{fig:eval_normal_assumption}
\end{figure}

To evaluate the modeling assumption that batch data are multivariate normal distributed, 
the distribution of residual data is compared to the estimated normal distribution. 
In \Fig\ref{fig:eval_normal_assumption}, samples from the four residuals are plotted pairwise 
against each other together with an ellipse representing the estimated covariance with 95\% 
confidence interval. Each column represents one batch of data and each row represent one combination of 
residual outputs. The blue ellipses represent the covariance estimated from all samples. 
To avoid an overestimation of the covariance due to outliers, the orange ellipses show the 
estimated covariances after removing 10\% of the outliers in each batch. These two approaches of estimating the pdfs 
of each batch will be further discussed in later sections. It is visible that the assumption that data is 
multivariate normal distribution is an approximation, especially when there are model uncertainties 
and outliers in residual data. Still, it gives some information about data distribution and correlation 
between features which can be used for fault classification. 

Before designing the set of $\mathcal{D}_j$ classifiers, the set of estimated pdfs 
for each fault class is then randomly split into a training and validation 
set where $67\%$ are used for training. The training set from each fault class 
is used to model each fault mode $\hat{\Omega}_i$ for the different fault 
classes represented in training data, see Table~\ref{tab:fault_modes}. The training 
set covers different realizations and magnitudes of each fault. Note that pdfs 
estimated from the fault-free case are included in all fault modes, i.e., 
$\hat{\Omega}_{NF} \subseteq \hat{\Omega}_{j}$ for all $j = 1,2, \ldots, m$. 

\subsection{Evaluating Fault Diagnosis Performance}

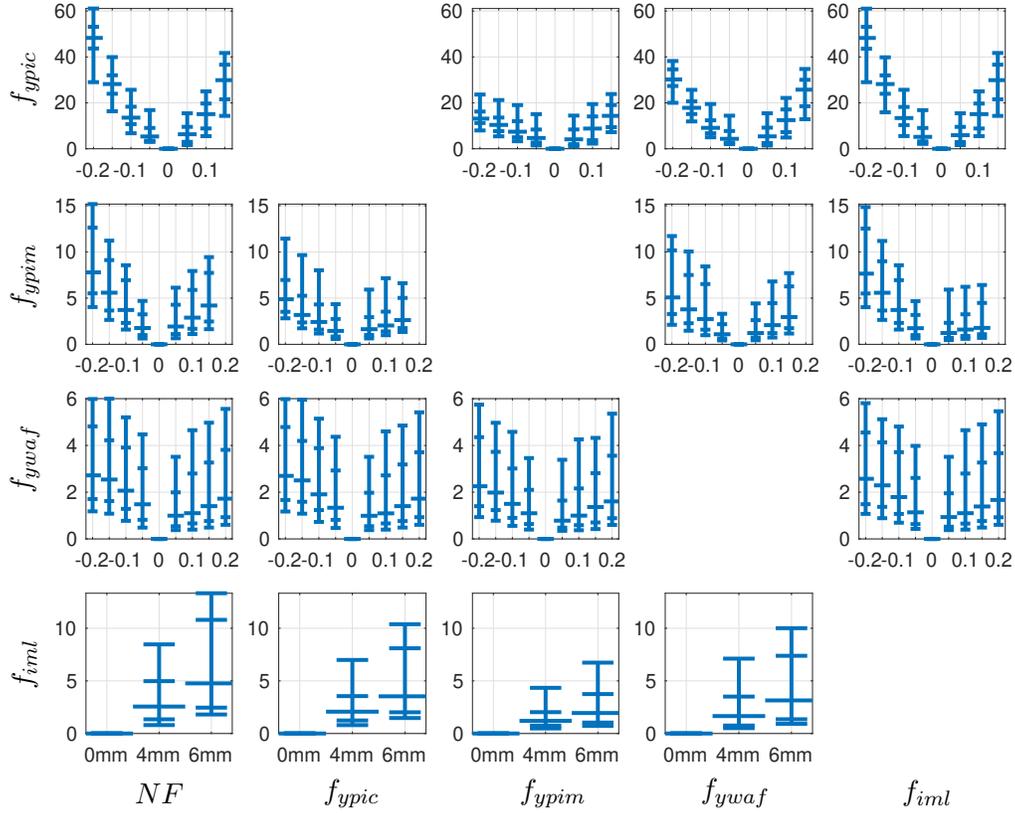
\begin{figure}
\centering
  \begin{tikzpicture}
    \node[anchor=south west,inner sep=0] (image) at (0,0) {
      \resizebox{0.93\columnwidth}{!}{\includegraphics{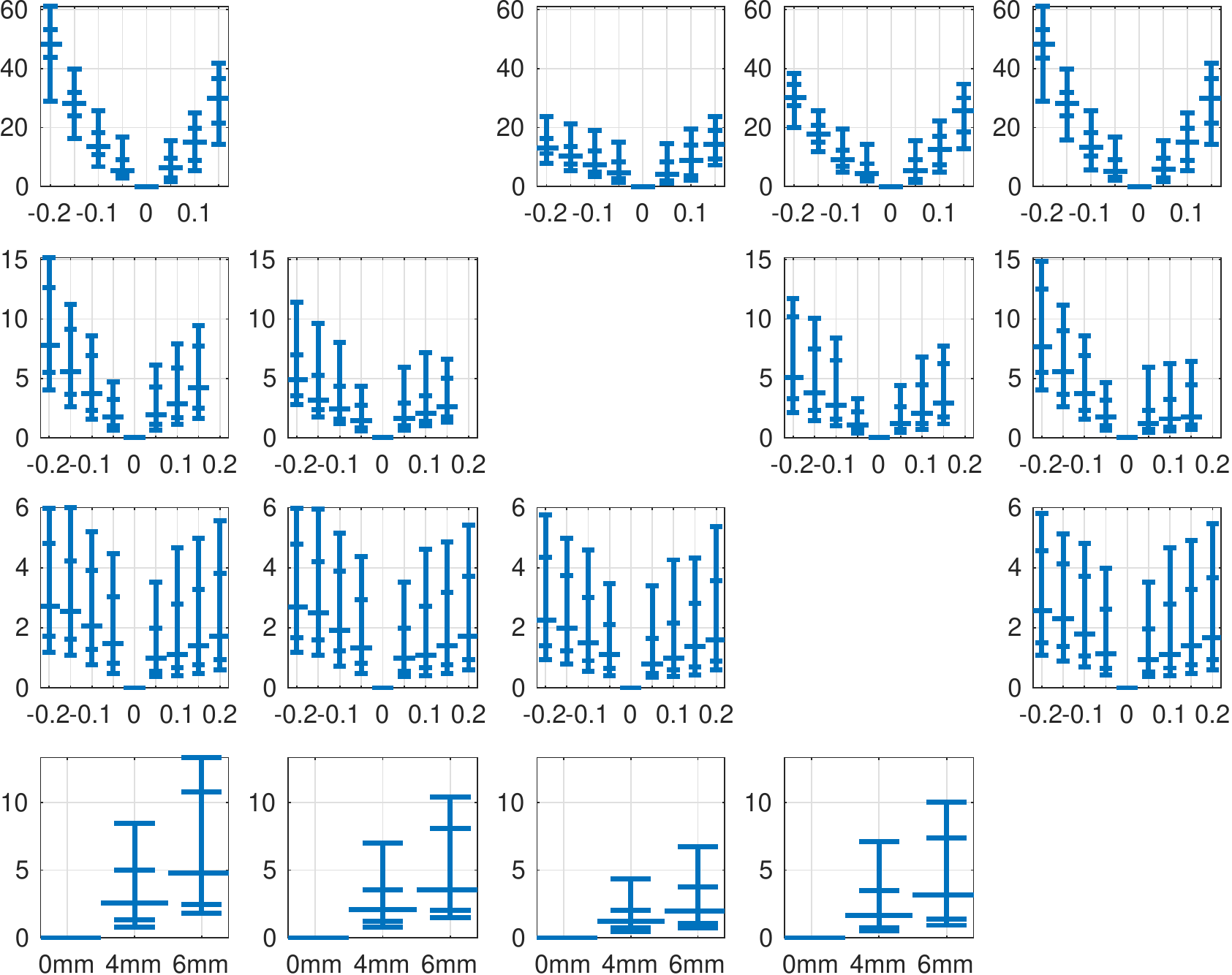}}
    };
    \begin{scope}[x={(image.south east)},y={(image.north west)}]
    \draw (-0.03,0.90) node{\rotatebox{90}{\small $f_{ypic}$}};
    \draw (-0.03,0.64) node{\rotatebox{90}{\small $f_{ypim}$}};
    \draw (-0.03,0.40) node{\rotatebox{90}{\small $f_{ywaf}$}};
    \draw (-0.03,0.13) node{\rotatebox{90}{\small $f_{iml}$}};
    \draw (0.11,-0.04) node{\small $NF$};        
    \draw (0.31,-0.04) node{\small $f_{ypic}$};        
    \draw (0.52,-0.04) node{\small $f_{ypim}$};        
    \draw (0.71,-0.04) node{\small $f_{ywaf}$};     
    \draw (0.91,-0.04) node{\small $f_{iml}$};     
    \end{scope}
  \end{tikzpicture}
	\caption{Evaluating the distinguishability measure \eqref{eq:dij} between fault modes (modeled using all data) as function of fault size. Each vertical line shows the quantiles $\{10\%, 25\%, 50\%, 75\%, 90\%\}$ of the distribution of distinguishability measures. Each plot $(i,j)$ shows that fault $f_i$ is easier to distinguish from fault mode $f_j$ with increasing fault size which is expected. Note that the distinguishability measure is non-symmetric, e.g., it is easier to distinguish $f_{ypic}$ from $f_{ypim}$ (or $f_{ywaf}$) than vice versa.}
\label{fig:dij_faults}
\end{figure}

The first step of the analysis is to evaluate the set of modeled fault modes to quantify 
how easy it is to distinguish between the different fault classes. In the analysis of fault diagnosis 
performance using the distinguishability measure, all available datasets from the different fault 
classes are used. The distinguishability measure 
is evaluated for all pdfs $p \in \hat{\Omega}_i$ with respect to all other fault modes and 
the distributions of $\mathcal{D}_{i,j}(p)$ values for different fault magnitudes when the batch size $N = 100$ are plotted in 
\Fig\ref{fig:dij_faults}. The subplot at position $(i,j)$ shows the distribution of $\mathcal{D}_{i,j}(p)$ 
for all $p \in \hat{\Omega}_i$. The marks on each vertical line represent the 
$10\%$, $25\%$, $50\%$, $75\%$, and $90\%$ quantiles. 
The results show that detection and isolation performance, in general, improves with 
increasing fault magnitude and that all faults are distinguishable from 
each other. In addition, fault $f_{ypic}$ should be the easiest of 
the three sensor faults to distinguish while, e.g., $f_{ywaf}$ is more difficult since the 
distinguishability measure is significantly smaller. Another observation is that the 
distinguishability measure is not symmetric, i.e., it might not be as easy 
to distinguish mode $f_i$ from $f_j$ as the other way around \citep{eriksson2013method}. For example, it is easier 
to distinguish $f_{ypic}$ from $f_{ypim}$ than vice versa which is shown by that the distinguishability 
measure is larger, see \Fig\ref{fig:dij_faults}. Also, it is easier to distinguish each fault mode 
from the fault-free mode, see the leftmost column in \Fig\ref{fig:dij_faults}, than to distinguish from the other 
fault modes, which is consistent with \eqref{eq:dij_det_isol}. 

Some of the results are summarized in Table~\ref{tab:Dij_varying_batch} corresponding to 
detection performance of the different fault classes. The tables show the mean value of 
$\mathcal{D}_{i,j}(p)$ when $p$ are estimated from batches of different sizes 
(between 50 and 300 samples).  It is visible that $\mathcal{D}_{i,j}(p)$ increases 
for fault sizes with higher distinguishability values when increasing the batch size, while it is stable 
or slightly decreasing when the distinguishability measure is small. In general, longer batch sizes  
make it easier to distinguish fault $f_{ypic}$ while it becomes more difficult for the other three fault 
classes. One explanation is that shorter batches can make it easier to distinguish the impact of the 
fault when fault excitation and residual noise level varies over time. 

{
\setlength{\tabcolsep}{3pt}
\begin{table}[h!]
\caption{Comparison of mean values of distinguishability measure of fault detection as a function of fault size and used batch size when estimating pdfs.}
\label{tab:Dij_varying_batch}
\centering
\footnotesize
\begin{tabular}{ c | c c c c }
\multicolumn{5}{c}{$\mathcal{D}_{fypic,NF}$} \\
\hline
$\theta$ \textbackslash $N$ & 50 & 100 & 200 & 300 \\
\hline
  -20\% & 37.7 & 45.9 & 69.5 & 77.1 \\
  -15\% & 23.7 & 27.8 & 39.7 & 43.0 \\
  -10\% & 13.4 & 14.8 & 18.6 & 19.1 \\  
    -5\% & 7.7   & 7.2 & 7.0 & 6.6 \\ 
     5\% & 9.1   & 7.1 & 7.0 & 7.3 \\ 
    10\%& 17.0 & 14.9 & 18.3 & 20.9 \\ 
    15\%& 31.0 & 28.8 & 40.0 & 45.6 \\     
\hline  
\end{tabular}\quad\quad\quad%
\begin{tabular}{ c | c c c c }
\multicolumn{5}{c}{$\mathcal{D}_{fypim,NF}$} \\
\hline
$\theta$ \textbackslash $N$ & 50 & 100 & 200 & 300 \\
\hline
  -20\% & 9.5 & 9.4 &10.1 & 10.7 \\
  -15\% & 6.7 & 6.5 & 6.8 & 7.1 \\
  -10\% & 5.1 & 4.6 & 4.2 & 4.2 \\  
    -5\% & 2.6 & 2.2 & 1.8 & 1.6\\ 
     5\% & 3.2 & 2.8 & 1.7 & 2.3 \\ 
    10\%& 4.7 & 4.2 & 3.9 & 3.8 \\ 
    15\%& 6.3 & 6.0 & 6.0 & 6.1 \\     
\hline  
\end{tabular}

\vspace{0.3cm}

\begin{tabular}{ c | c c c c }
\multicolumn{5}{c}{$\mathcal{D}_{fywaf,NF}$} \\
\hline
$\theta$ \textbackslash $N$ & 50 & 100 & 200 & 300 \\
\hline
  -20\% & 3.9 & 3.3 & 3.0 & 2.9 \\
  -15\% & 3.7 & 3.1 & 2.6 & 2.4 \\
  -10\% & 3.1 & 2.7 & 2.1 & 1.9 \\  
    -5\% & 2.4 & 2.0 & 1.8 & 1.6 \\ 
     5\% & 1.8 & 1.4 & 1.1 & 0.8 \\ 
    10\%& 2.2 & 1.9 & 1.4 & 1.1 \\ 
    15\%& 2.6 & 2.1 & 1.6 & 1.4 \\     
    20\%& 3.0 & 2.5 & 2.1 & 2.0 \\     
\hline  
\end{tabular}\quad\quad\quad%
\begin{tabular}{ c | c c c c }
\multicolumn{5}{c}{$\mathcal{D}_{fiml,NF}$} \\
\hline
$\theta$ \textbackslash $N$ & 50 & 100 & 200 & 300 \\
\hline
    4mm& 4.0 & 3.6 & 3.0 & 2.6 \\     
    6mm& 6.9 & 6.7 & 6.4 & 5.8 \\     
\hline  
\end{tabular}
\end{table}
}
\subsection{Fault Classification}

The open set fault classification algorithm described in Section~\ref{sec:classification}
is implemented where a $\mathcal{D}_j$ classifier is trained for each 
known fault class $f_j$, as described in Section~\ref{sec:classification}. 
A threshold is selected based 
on the distribution of the within-class distinguishability measure \eqref{eq:within_class_dij} 
using kernel density estimation to have a $5\%$ outlier rate. The calibrated thresholds for each fault class 
are presented in Table~\ref{tab:J}. For comparison, the kernel density estimations of both training 
and validation data for one fault class are shown in \Fig\ref{fig:ksdensity_J}.

\begin{table}[h!]
\caption{Example of calibrated thresholds $J_j$ for each $\mathcal{D}_j$ classifier.}
\label{tab:J}
\centering
\begin{tabular}{ c c c c c }
\hline
$J_{NF}$ & $J_{fypic}$ & $J_{fypim}$ & $J_{fywaf}$ & $J_{fiml}$ \\
\hline
  1.98 & 3.03 & 2.43 & 2.36 & 2.31 \\
\hline  
\end{tabular}
\end{table}

\subsubsection{Classification of known fault classes}

\begin{figure}
\centering
  \begin{tikzpicture}
    \node[anchor=south west,inner sep=0] (image) at (0,0) {
      \resizebox{0.93\columnwidth}{!}{\includegraphics{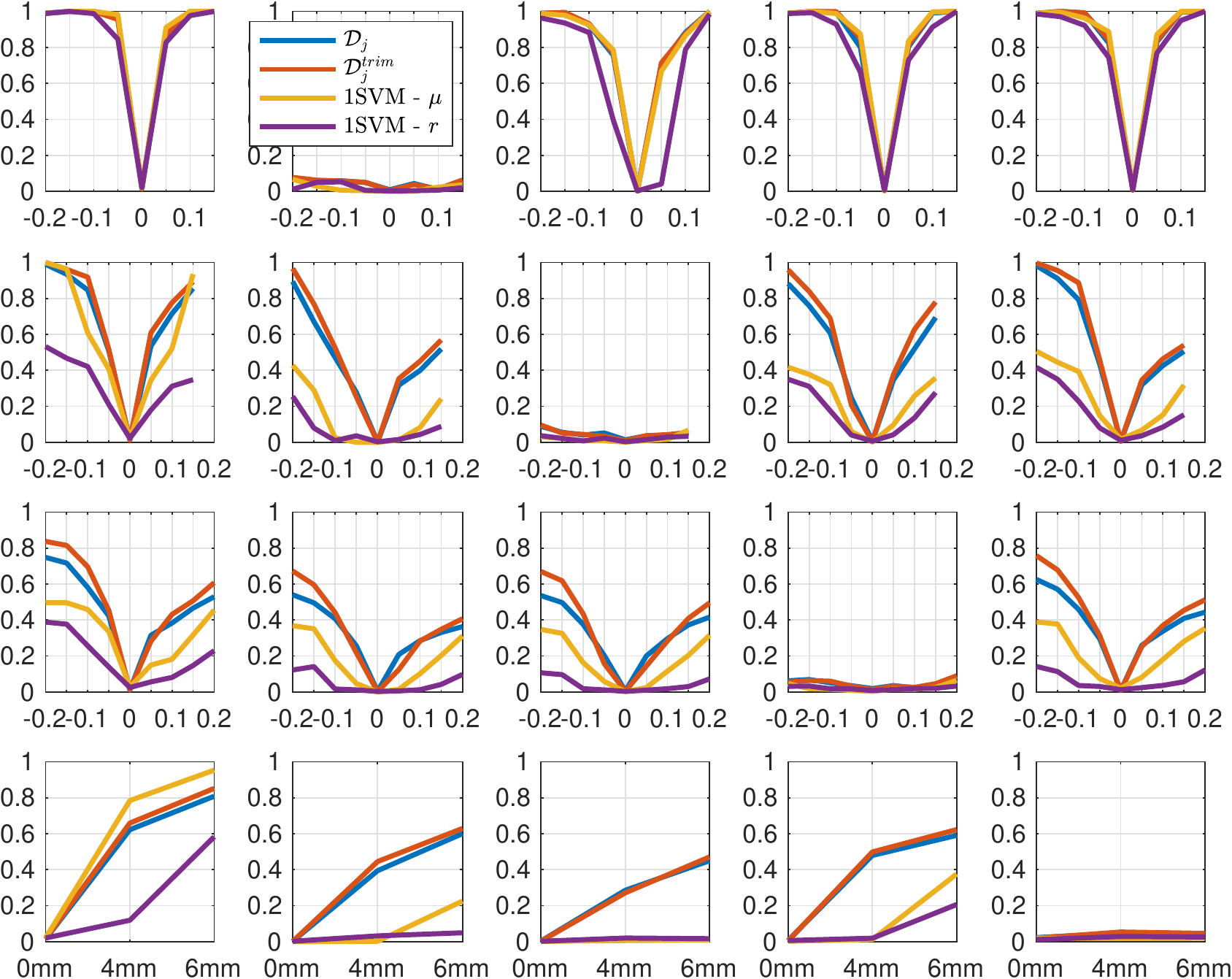}}
    };
    \begin{scope}[x={(image.south east)},y={(image.north west)}]
    \draw (-0.03,0.90) node{\rotatebox{90}{\small $f_{ypic}$}};
    \draw (-0.03,0.64) node{\rotatebox{90}{\small $f_{ypim}$}};
    \draw (-0.03,0.40) node{\rotatebox{90}{\small $f_{ywaf}$}};
    \draw (-0.03,0.13) node{\rotatebox{90}{\small $f_{iml}$}};
    \draw (0.11,-0.04) node{\small $NF$};        
    \draw (0.31,-0.04) node{\small $f_{ypic}$};        
    \draw (0.52,-0.04) node{\small $f_{ypim}$};        
    \draw (0.71,-0.04) node{\small $f_{ywaf}$};     
    \draw (0.91,-0.04) node{\small $f_{iml}$};       
    \end{scope}
  \end{tikzpicture}
	\caption{Monte Carlo evaluation of detection and isolation performance using a batch size of 50 samples. Each curve shows the average of 10 runs. The decision boundary is selected using an approximate $ 2 \% $ training outlier rate. Each subplot at position $(i,j)$ shows the probability of rejecting fault class $f_j$ when a fault $f_i$ occurs as function of fault size. The $\mathcal{D}_j$ classifier using trimmed covariance estimates is labelled $\mathcal{D}_j^{trim}$.} 
	\label{fig:classification_50}
\end{figure}
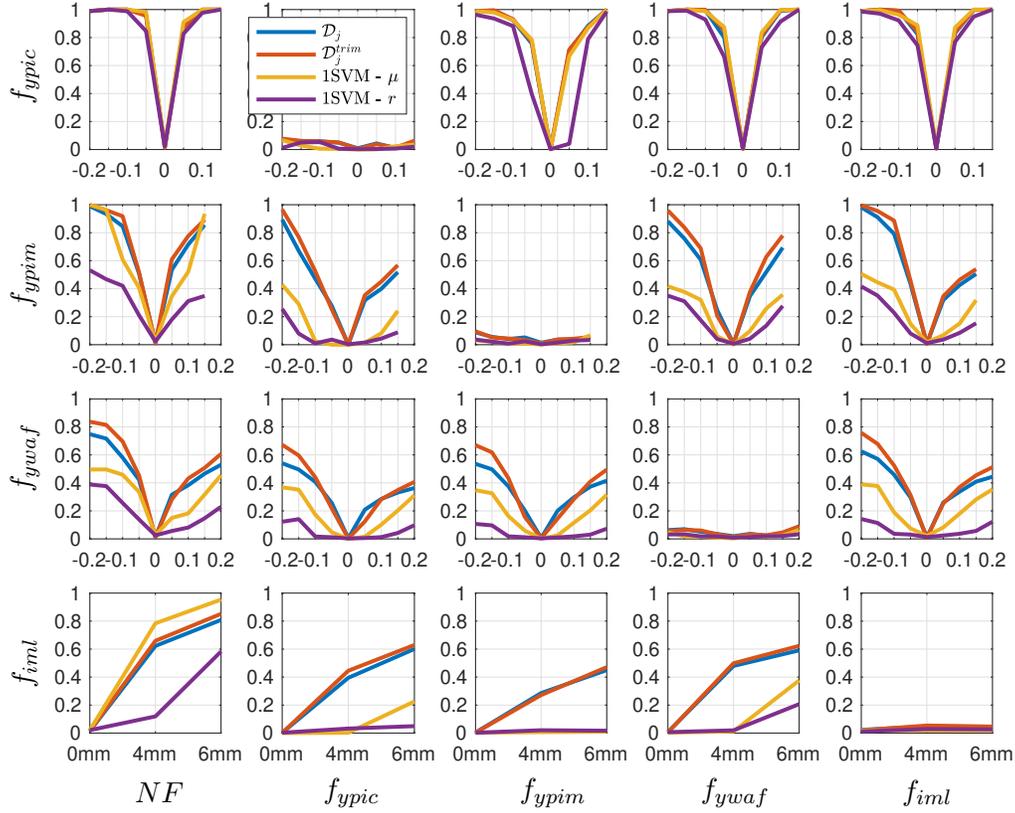

\begin{figure}
\centering
  \begin{tikzpicture}
    \node[anchor=south west,inner sep=0] (image) at (0,0) {
      \resizebox{0.93\columnwidth}{!}{\includegraphics{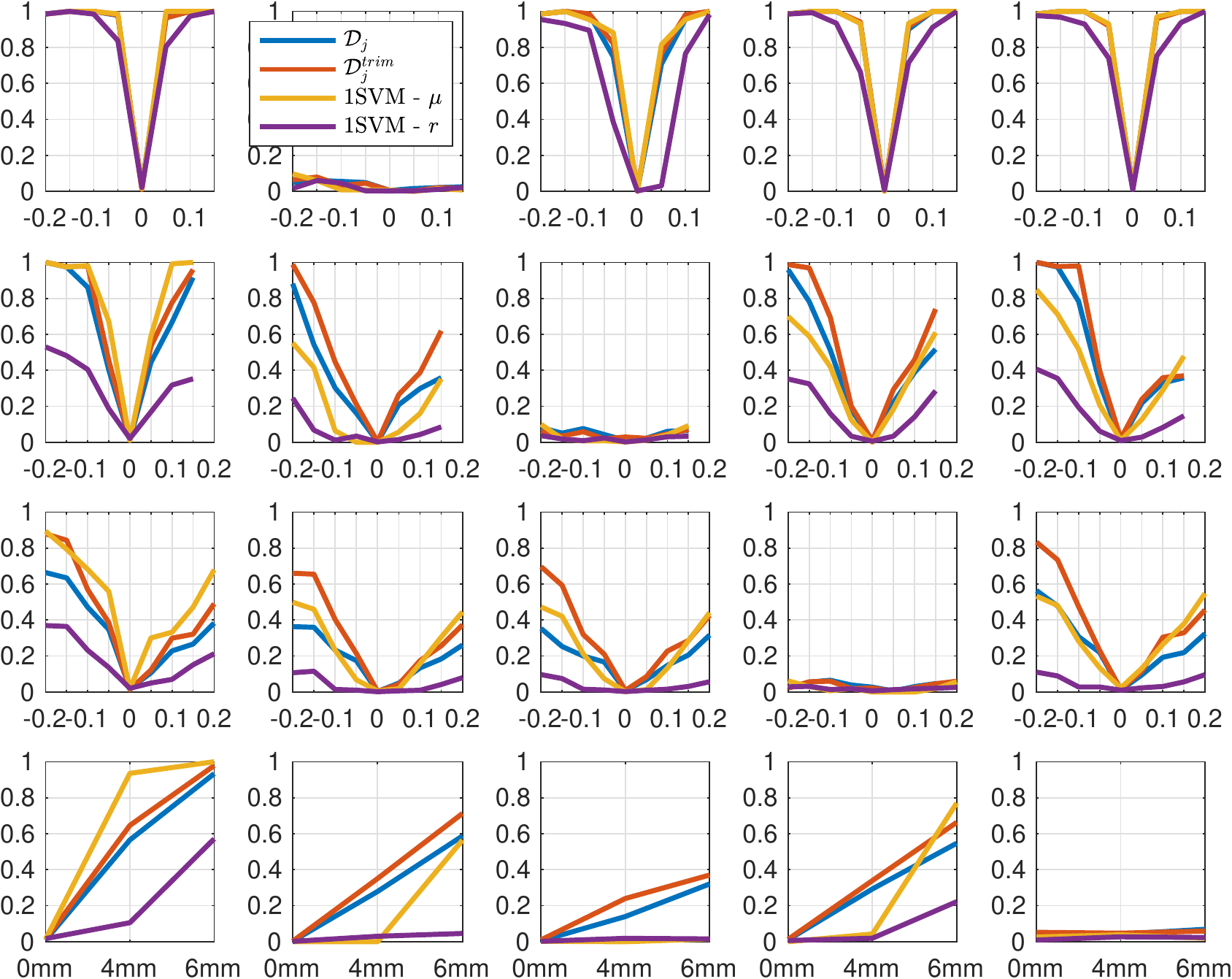}}
    };
    \begin{scope}[x={(image.south east)},y={(image.north west)}]
    \draw (-0.03,0.90) node{\rotatebox{90}{\small $f_{ypic}$}};
    \draw (-0.03,0.64) node{\rotatebox{90}{\small $f_{ypim}$}};
    \draw (-0.03,0.40) node{\rotatebox{90}{\small $f_{ywaf}$}};
    \draw (-0.03,0.13) node{\rotatebox{90}{\small $f_{iml}$}};
    \draw (0.11,-0.04) node{\small $NF$};        
    \draw (0.31,-0.04) node{\small $f_{ypic}$};        
    \draw (0.52,-0.04) node{\small $f_{ypim}$};        
    \draw (0.71,-0.04) node{\small $f_{ywaf}$};     
    \draw (0.91,-0.04) node{\small $f_{iml}$};       
    \end{scope}
  \end{tikzpicture}
	\caption{Monte Carlo evaluation of detection and isolation performance using a batch size of 300 samples. Each curve shows the average of 10 runs. The decision boundary is selected using an approximate $ 2 \% $ training outlier rate. Each subplot at position $(i,j)$ shows the probability of rejecting fault class $f_j$ when a fault $f_i$ occurs as function of fault size.}
	\label{fig:classification_300}
\end{figure}

First, the set of $\mathcal{D}_j$ classifiers are evaluated using pdfs 
from the known fault classes in the validation set. Since each batch of residual data is modeled as a multivariate normal distribution,
estimation of the covariance matrix can be sensitive to outliers in the residual outputs. Therefore, an additional set of 
$\mathcal{D}_j$ classifiers are trained using trimmed estimates of the covariance matrix by first removing 10\% of 
the outliers, denoted $\mathcal{D}_j^{trim}$. For comparison, 
two sets of one-class support vector machines (1SVM) \cite{scholkopf2000support} are trained. 
The 1SVM classifiers are implemented using the function \texttt{fitcsvm} in \texttt{Matlab} and 
their kernel parameters are fit to training data using a subsampling heuristic \cite{matlab2018b}. 
When analyzing the results of the $\mathcal{D}_j$ classifier, the false alarm 
rate ($<2\%$) was lower compared to the outlier rate of $5\%$ when selecting the threshold $J_j$. 
Therefore, an outlier rate of 2\% was selected when training the 1SVM classifiers to give more 
comparable results. The first 1SVM classifier uses the mean of the pdfs as input, referred to as 
1SVM-$\mu$ and the second set uses the raw residual data as input, referred to as 1SVM-r. 
Results are presented when the pdfs are estimated 
using two different batch sizes: 50 samples and 300 samples. Ideally, the probability of 
rejecting a fault class should be small for the true class and large for all other fault classes. 
The probabilities of rejecting each fault class given data from different fault 
realizations using the three different open set fault classifiers are shown in \Fig\ref{fig:classification_50} 
using batch size 50 and \Fig\ref{fig:classification_300} using batch size 300, respectively. 
The curves show the mean values of 10 Monte Carlo evaluations. 

Ideally, probability of rejection should be 100\%, for all non-zero fault sizes except when 
classifying data from the same fault class that the classifier has been trained on.  
In that case, probability of rejection should be as small as possible because this would otherwise 
mean that the true fault class is rejected. 
Classification performance is consistent with the previous results in \Fig\ref{fig:dij_faults}
showing $f_{ypic}$ is easiest to diagnose, since the probabilities to reject the wrong fault hypotheses 
are higher compared to the other fault scenarios. The most difficult fault to diagnose is $f_{ywaf}$. 
The results in Table~\ref{tab:Dij_varying_batch} are also consistent with the analysis in 
\Fig\ref{fig:dij_faults} since classification performance of the $\mathcal{D}_j$ classifier 
is better for smaller batch sizes. 

When comparing the results between the 
four algorithms in \Fig\ref{fig:classification_50} and \Fig\ref{fig:classification_300}, the 1SVM-r 
classifier has the overall worst performance while 1SVM-$\mu$ improves with increasing batch size. 
However, using trimmed estimates of the covariance matrix in the $\mathcal{D}_j^{trim}$ classifiers improve
classification performance for both shorter and longer batch sizes. Table~\ref{tab:classifiers_varying_batch} 
shows the detection performance values to compare $\mathcal{D}_j$, $\mathcal{D}_j^{trim}$, and 1SVM-$\mu$, 
for the two different batch sizes. These results indicate that outliers in residual data could explain the reduced 
performance of the original $\mathcal{D}_j$ classifiers for longer batch sizes. Thus, one solution to improve 
classification performance is to apply robust estimation of the normal distribution parameters. 

{
\setlength{\tabcolsep}{1.5pt}
\begin{table}[h!]
\caption{Results from comparing fault detection accuracy in \Fig\ref{fig:classification_50} and \Fig\ref{fig:classification_300} of $\mathcal{D}_j$ classifier, $\mathcal{D}_j^{trim}$, and 1SVM-$\mu$, respectively, when a fault is present, and false alarm rate when fault size is $0\%$.}
\label{tab:classifiers_varying_batch}
\centering
\tiny
\begin{tabular}{ c | c c c | c c c}
$f_{ypic}$ & \multicolumn{3}{c}{ batch size 50} & \multicolumn{3}{c}{ batch size 300} \\
\hline
$\theta$ & $\mathcal{D}_j$ & $\mathcal{D}_{j}^{trim}$ & 1SVM-$\mu$ & $\mathcal{D}_j$ & $\mathcal{D}_{j}^{trim}$ & 1SVM-$\mu$\\
\hline
  -20\% & 99.0\% & 99.0\%& 99.0\% & 98.6\% & 98.6\% & 98.6\% \\
  -15\% & 100\% & 100\%& 100\% & 100\% & 100\% &100\%\\
  -10\% & 100\% & 100\%& 100\% & 100\% & 100\% &100\% \\  
    -5\% & 96.1\%  & 95.5\%& 98.2\% & 97.6\% & 98.1\% & 98.6\%\\ 
     0\% & 1.5\% & 1.4\%& 2.0\% & 2.1\% & 1.1\% & 1.4\% \\
     5\% & 87.4\%  & 86.9\%& 91.0\% & 96.7\% & 96.4\% & 100\% \\ 
    10\%& 100\% & 100\%& 100\%  & 100\%& 100\% & 100\%\\ 
    15\%& 100\% & 100\% & 100\%  & 100\%& 100\% & 100\%\\     
\hline  
\end{tabular} \hfill% 
\begin{tabular}{ c | c c c | c c c}
$f_{ypim}$ & \multicolumn{3}{c}{ batch size 50} & \multicolumn{3}{c}{ batch size 300} \\
\hline
$\theta$ & $\mathcal{D}_j$ & $\mathcal{D}_{j}^{trim}$ & 1SVM-$\mu$ & $\mathcal{D}_j$ & $\mathcal{D}_{j}^{trim}$ & 1SVM-$\mu$\\
\hline
  -20\% & 98.9\% & 100\%& 100\% & 100\% & 100\% & 100\%\\
  -15\% & 93.4\% & 96.1\%& 96.3\% & 97.4\% & 97.6\% & 97.4\%\\
  -10\% & 84.6\% & 91.8\%& 60.8\% & 85.2\% & 98.0\% & 98.2\%\\  
    -5\% & 50.8\%  & 51.7\%& 40.6\% & 38.4\% & 45.3\% & 67.1\%\\ 
     0\% & 1.3\% & 1.2\%& 2.2\% & 1.5\% & 1.3\% & 1.1\%\\
     5\% & 53.5\%  & 60.9\% & 34.6\% & 42.1\% & 53.4\% & 59.0\%\\ 
    10\%& 71.6\% & 77.6\% & 52.1\%  & 65.0\% & 77.6\% & 99.2\%\\ 
    15\%& 85.5\% & 89.3\% & 93.5\%  & 90.8\% & 96.0\% & 100\%\\     
\hline  
\end{tabular}

\vspace{0.3cm}

\begin{tabular}{ c | c c c | c c c}
$f_{ywaf}$ & \multicolumn{3}{c}{ batch size 50} & \multicolumn{3}{c}{ batch size 300} \\
\hline
$\theta$ & $\mathcal{D}_j$ & $\mathcal{D}_{j}^{trim}$  & 1SVM-$\mu$ & $\mathcal{D}_j$ & $\mathcal{D}_j^{trim} $ & 1SVM-$\mu$\\
\hline
  -20\% & 74.8\% & 83.7\%& 49.6\% & 63.7\%& 87.9\% & 89.4\%\\
  -15\% & 71.6\% & 81.5\%& 49.6\% & 59.5\%& 84.3\% & 79.3\%\\
  -10\% & 58.1\% & 69.6\%& 45.8\% & 44.9\%& 56.9\% & 68.1\%\\  
    -5\% & 42.2\%  & 45.3\%& 33.6\% & 33.7\%& 38.6\% & 55.9\%\\ 
     0\% & 1.1\% & 1.3\%& 2.8\% & 2.0\% & 1.4\%& 1.7\%\\
     5\% & 31.6\%  & 28.3\%& 15.0\% & 10.7\%& 12.2\%&30.1\%\\ 
    10\%& 38.3\% & 43.1\%& 18.2\%  & 21.6\%& 30.0\%& 33.2\%\\ 
    15\%& 46.6\% & 50.7\%& 31.6\%  & 25.5\%& 32.1\%&47.2\%\\ 
    20\%& 53.0\% & 60.7\%& 45.5\%  & 36.7\%& 49.0\%& 67.9\%\\     
\hline  
\end{tabular}\hfill%
\begin{tabular}{ c | c c c | c c c}
$f_{iml}$ & \multicolumn{3}{c}{ batch size 50} & \multicolumn{3}{c}{ batch size 300} \\
\hline
$\theta$ & $\mathcal{D}_j$ & $\mathcal{D}_{j}^{trim}$ & 1SVM-$\mu$ & $\mathcal{D}_j$ & $\mathcal{D}_{j}^{trim}$ & 1SVM-$\mu$\\
\hline
  0mm & 1.5\% & 1.5\%& 2.1\% & 0.9\% & 1.4\% & 1.4\% \\
  4mm & 62.3\% & 65.9\% & 78.4\% & 53.6\% & 64.6\% & 93.6\%\\
  6mm & 80.8\% & 85.2\%& 95.3\% & 93.1\% & 97.8\% & 100\% \\  
\hline  
\end{tabular}
\end{table}
}

Another interesting observation is that $\mathcal{D}_j$ and $\mathcal{D}_j^{trim}$ 
are significantly better at distinguishing fault $f_{iml}$ from $f_{pim}$ compared to the 1SVM classifiers. 
These faults are expected to be difficult to distinguish from each other since data from the two classes are overlapping, 
as illustrated in Figure~\ref{fig:residual_faults}. However, classifying distributions instead of only sample means, makes it 
possible to distinguish between the two classes, which is expected from the quantitative analysis in \Fig~\ref{fig:dij_faults}. 
The results from the experiments show that approximating residual data using a multivariate normal distribution is sufficient 
in this case study to distinguish between the different fault classes. However, it is likely that classification accuracy can be 
improved by using a more flexible model to estimate the distribution, at the expense of increased computational cost. Another 
approach to improve classification accuracy, especially detection of small faults, is to compare the diagnosis output over consecutive 
batches. Since more and more data will be collected over time, the classification accuracy in Figure~\ref{fig:classification_50} 
and Figure~\ref{fig:classification_300}, respectively, can most likely be improved by allowing a longer time to detect. 

An advantage of using the $\mathcal{D}_j$ classifiers is that less memory is needed to store 
information, since it is sufficient to store distribution parameters and not the original batch data. 
If different remote diagnosis solutions are used which have access to more computational 
capabilities for data analysis compared to what is available in an on-board diagnosis system, 
it is also relevant to minimize the amount of transmitted data \cite{song2005remote,langarica2019industrial}. 

A limitation of the $\mathcal{D}_j$ classifier is that evaluating \eqref{eq:dij} corresponds to a nearest 
neighbor search problem, which can be computationally heavy when the cardinality of the set $\hat{\Omega}_j$ 
is large. Parallelization and the use of different heuristics could help 
to prune the search space and thus, significantly reduce the computation time, see for example \cite{cayton2008fast}.  

\subsubsection{Classification of unknown fault class}

Fault classification is here performed by rejecting fault hypotheses, as described in 
Section~\ref{sec:computing_fault_hypotheses} where each $\mathcal{D}_j$~classifier 
is used to determine if fault class $f_j$ can explain the distribution of batch data or not. 
Note that each pdf in the validation set is evaluated against each known fault class, 
independently of the other fault classes. Thus, the performance of classifying 
unknown fault classes is based on the probabilities that all known fault classes are rejected.
This can be evaluated by using the previous results in Figures~\ref{fig:classification_50} and 
\ref{fig:classification_300}. A scenario where one of the fault classes is assumed 
to be unknown can be evaluated by ignoring the results in the column that corresponds to the 
$\mathcal{D}_j$~classifier which models the unknown fault class. 

In the first example, an unknown fault scenario is evaluated using validation data from fault $f_{ypic}$, i.e., 
it is assumed that there are no training data from that fault. 
To evaluate the probability that the known fault classes, $NF$, $f_{ypim}$, and $f_{ywaf}$, 
are correctly rejected in this scenario, the first row is analyzed in 
\Fig\ref{fig:classification_50} (and \Fig\ref{fig:classification_300}) while ignoring the second 
column which corresponds to the $\mathcal{D}_j$ classifier modeling fault $f_{ypic}$. In this 
case all classifiers have more than $80\%$ probability of rejecting the $NF$ class for all realizations 
of $f_{ypic}$ in validation data. Similarly, the other known fault classes have a high probability 
of correctly being rejected, except for the 1SVM-$r$ classifier modeling fault $f_{ypim}$ which has a low 
probability of rejecting $f_{ypim}$ for scenarios when $f_{ypic}$ is of size $5\%$. This shows 
that $f_{ypic}$ is likely to be correctly classified as an unknown fault since all known fault classes 
will be rejected with high probability. Note that the probabilities of rejecting each fault class are
evaluated by classifying one pdf. The probability of correctly classifying smaller faults can be 
significantly improved by evaluating multiple pdfs corresponding to consecutive batches and 
compare the rate that each fault class is rejected with respect to the probability that it is falsely 
rejected when it is the true fault class. 

As a second example, fault $f_{iml}$ is simulated as an unknown fault while the other fault classes 
in the training set are known. Then, the last row is 
analyzed in \Fig\ref{fig:classification_50} (and \Fig\ref{fig:classification_300}) while ignoring the 
last column. When comparing the probability of correctly rejecting each known fault class in the last row
the 1SVM-$\mu$ performs better than the $\mathcal{D}_j(p)$ and $\mathcal{D}_j(p)^{trim}$ 
classifiers to detect the fault (rejecting the $NF$ class). The 1SVM-$r$ classifier modeling the $NF$ 
class has significantly worse performance, especially for the 4mm leakage. 
However, when analyzing the probability of rejecting the other known fault classes, $f_{ypic}$, $f_{ypim}$, 
and $f_{ywaf}$, the $\mathcal{D}_j(p)$ and $\mathcal{D}_j(p)^{trim}$ classifiers have an overall
better performance compared to 1SVM-$\mu$ and 1SVM-$r$, especially for the 4mm leakage. 
The 1SVM-$r$ is not able to reject $f_{ypic}$ and $f_{ypim}$, for any of the leakage sizes 
since the probabilities of correctly rejecting those fault classes are not significantly higher than the probability
that each classifier is falsely rejecting the true fault class. For the larger batch size, the 1SVM-$\mu$
performs significantly better for the 6mm leakage compared to the 4mm leakage but is not able to 
reject $f_{ypim}$. 

The two examples show the ability of the proposed open set fault classification algorithm, using 
$\mathcal{D}_j(p)$ or $\mathcal{D}_j(p)^{trim}$ classifiers, to identify unknown fault classes. The 
examples also illustrate the computational benefits since there is no need of recalibrating 
the whole set of $\mathcal{D}_j$ classifiers for the existing set of known fault classes when 
updating one classifier $\mathcal{D}_j$ with new training data or when including a new classifier 
for a new fault class. 

\subsection{Fault Size Estimation}

The fault estimation algorithm \eqref{eq:lambda_est} described in Section~\ref{sec:estimation} 
is applied to the validation set when the corresponding fault size of each pdf in the training set is known. The numerical KL divergence 
\eqref{eq:kld_mc} is evaluated using 1000 Monte Carlo samples where the cost function is minimized using the 
10 pdfs in training data with the smallest $\mathcal{D}_j(p)$ values. The fault size estimation for each pdf $p$ using 
\eqref{eq:lambda_est} is denoted $\hat{\theta}_{KL}$. 

The prediction results 
for each fault class are shown in Table \ref{tab:theta_estimation} by presenting the 10\% and 90\% quantiles 
of the fault size predictions. To evaluate the proposed data-driven fault size estimation algorithm \eqref{eq:lambda_est}, 
it is compared to estimating the fault size $\hat{\theta}_{mean}$ by computing the mean fault size of the 10 pdfs with 
the smallest $\mathcal{D}_j(p)$ values. For the proposed algorithm, the true fault sizes are within the intervals
while for the estimate $\hat{\theta}_{mean}$, the true fault sizes are outside the intervals for the largest realizations of $f_{ywaf}$. 
The algorithm \eqref{eq:lambda_est}, in general, has a narrower interval compared to using the mean estimate. 
Note that a limitation of the evaluated methods is that they are not capable of estimating fault sizes beyond 
what is available in training data.   

The prediction error correlates with the analysis of distinguishability between 
different fault modes in \Fig\ref{fig:dij_faults}. The intervals are smaller for $f_{ypic}$ 
while $f_{ywaf}$ has the largest intervals. One solution to improve the estimation 
accuracy over time is to look at the distribution of the estimated fault size over consecutive 
batches.

\setlength{\tabcolsep}{1.5pt}
\begin{table}[h!]
\caption{Results from using the fault size estimation algorithm \eqref{eq:lambda_est} on engine residual data. For each fault size $\theta$, the intervals represent the 10\% and 90\% quantiles of the estimates $\hat{\theta}_{KL}$ that are computed using \eqref{eq:KLD_minimization} and $\hat{\theta}_{mean}$ which is the mean fault size computed from the 10 pdfs with smallest $\mathcal{D}_j(p)$ values.}
\label{tab:theta_estimation}
\centering
{
\footnotesize
\begin{tabular}{ c | c | c}
\multicolumn{3}{c}{ $f_{ypic}$} \\
\hline
$\theta$ & $\hat{\theta}_{KL}$ & $\hat{\theta}_{mean}$ \\
\hline
  -20\% & $[-20.0\%, -18.4\%]$ & $[-20.0\%, -14.5\%]$ \\
  -15\% & $[-15.0\%, -14.8\%]$ & $[-15.5\%, -11.4\%]$ \\
  -10\% & $[-10.0\%, -10.0\%]$ & $[-10.0\%, -8.0\%]$ \\
    -5\% & $[-5.0\%, -5.0\%]$ & $[-5.5\%, -5.0\%]$ \\
     0\% & $[-1.4\%, 0.0\%]$ & $[-2.0\%, 0.5\%]$ \\
     5\% & $[5.0\%, 5.0\%]$ & $[4.5\%, 5.0\%]$ \\
    10\%& $[10.0\%, 10.0\%]$ & $[8.5\%, 10.0\%]$ \\
    15\%& $[15.0\%, 15.0\%]$ & $[13.0\%, 15.0\%]$ \\
\hline  
\end{tabular} \hfill% 
\begin{tabular}{ c | c | c}
\multicolumn{3}{c}{ $f_{ypim}$} \\
\hline
$\theta$ & $\hat{\theta}_{KL}$ & $\hat{\theta}_{mean}$ \\
\hline
  -20\% & $[-20.0\%, -17.2\%]$ & $[-20.0\%, -17.0\%]$ \\
  -15\% & $[-16.2\%, -12.3\%]$ & $[-16.5\%, -12.5\%]$ \\
  -10\% & $[-12.8\%, -8.4\%]$ & $[-12.5\%, -7.0\%]$ \\
    -5\% & $[-6.1\%, -1.4\%]$ & $[-7.2\%, -1.5\%]$ \\
     0\% & $[-3.2\%, 1.8\%]$ & $[-3.0\%, 2.0\%]$ \\
     5\% & $[3.2\%, 9.2\%]$ & $[2.1\%, 9.5\%]$ \\
    10\%& $[8.6\%, 12.6\%]$ & $[8.1\%, 12.0\%]$ \\
    15\%& $[10.8\%, 15.0\%]$ & $[10.0\%, 15.0\%]$ \\
\hline  
\end{tabular} 
\vspace{0.3cm}

\begin{tabular}{ c | c | c}
\multicolumn{3}{c}{ $f_{ywaf}$} \\
\hline
$\theta$ & $\hat{\theta}_{KL}$ & $\hat{\theta}_{mean}$ \\
\hline
  -20\% & $[-20.0\%, -12.0\%]$ & $[-19.5\%, -11.6\%]$ \\
  -15\% & $[-17.4\%, -11.3\%]$ & $[-17.5\%, -7.9\%]$ \\
  -10\% & $[-13.5\%, -6.3\%]$ & $[-14.0\%, -5.7\%]$ \\
    -5\% & $[-10.1\%, -0.1\%]$ & $[11.0\%, 1.7\%]$ \\
     0\% & $[-2.6\%, 7.9\%]$ & $[-2.0\%, 7.5\%]$ \\
     5\% & $[1.3\%, 10.3\%]$ & $[-1.5\%, 9.0\%]$ \\
    10\%& $[4.4\%, 13.5\%]$ & $[5.0\%, 13.0\%]$ \\
    15\%& $[7.9\%, 17.6\%]$ & $[7.0\%, 16.5\%]$ \\
    20\%& $[10.2\%, 20.0\%]$ & $[8.2\%, 18.3\%]$ \\    
\hline  
\end{tabular} 
}
\end{table}

\section{Concluding Remarks and Future Works}
\label{sec:conclusions}

A data-driven framework for fault diagnosis of technical systems and time-series data is 
proposed that can handle imbalanced training data and unknown faults. The KL divergence 
is used as a similarity measure when evaluating if new data can be explained by that class 
or not. An advantage of the proposed $\mathcal{D}_j$ classifier, with respect to other black-box models, 
is interpretability, where the quantitative performance analysis and modeling of different 
fault modes can give valuable insights about the nature of different faults, and the ability 
to integrate fault size estimation within the proposed framework. The open set fault 
classification algorithm consists of a set of one-class classifiers modeling each fault 
class which makes it possible to identify all plausible fault hypotheses including scenarios 
with likely unknown faults. Instead of sample-by-sample classification, the KL divergence is used to 
classify if a batch of data can be explained by a given fault class or not. Experiments using 
real datasets from an internal combustion engine test bench show that the proposed framework
can predict which faults that are easy to classify. They also show that the $\mathcal{D}_j$ classifiers 
can classify faults that were difficult to distinguish using the 1-SVM classifiers and that 
it is possible to give an accurate estimation of the fault size without the need of a 
parametric model of the fault. 

Estimating pdfs from training data becomes complicated when the number of 
features grows. For future works, the objective is to adapt the proposed 
methods for large-scale problems by using distributed fault diagnosis techniques
but also reduce computation complexity of, e.g. \eqref{eq:dij} when training data 
grows, by using different heuristics or systematic search algorithms.
Another interesting continuation of this work is to extend the proposed methods 
for applications in condition monitoring and prognostics to be used for predicting 
system degradation rate by using batch data to track changes in fault size.

%% The Appendices part is started with the command \appendix;
%% appendix sections are then done as normal sections
%% \appendix

%% \section{}
%% \label{}

%% If you have bibdatabase file and want bibtex to generate the
%% bibitems, please use
%%
\bibliographystyle{elsarticle-num} 
\bibliography{main}

%% else use the following coding to input the bibitems directly in the
%% TeX file.

\end{document}